\documentclass[10pt,twocolumn,letterpaper]{article}

\usepackage[pagenumbers]{cvpr} 

\usepackage{graphicx}
\usepackage{amsmath}
\usepackage{amssymb}
\usepackage{booktabs}
\usepackage{multirow}

\usepackage[pagebackref,breaklinks,colorlinks]{hyperref}

\usepackage[capitalize]{cleveref}
\crefname{section}{Sec.}{Secs.}
\Crefname{section}{Section}{Sections}
\Crefname{table}{Table}{Tables}
\crefname{table}{Tab.}{Tabs.}

\def\cvprPaperID{6070} 
\def\confName{CVPR}
\def\confYear{2022}

\begin{document}

\title{Explore Spatio-temporal Aggregation for Insubstantial Object Detection: Benchmark Dataset and Baseline}
\author{Kailai Zhou, Yibo Wang, Tao Lv, Yunqian Li, Linsen Chen, Qiu Shen\footnotemark[1], Xun Cao\footnotemark[1] \\
Nanjing University, Nanjing, China\\
{\tt\small \{calayzhou,ybwang,lvtao,lyq,linsen\_chen\}@smail.nju.edu.cn, \{shenqiu,caoxun\}@nju.edu.cn}}




\maketitle
\footnotetext[1]{Corresponding author.}
\begin{abstract}
 We endeavor on a rarely explored task named Insubstantial Object Detection (IOD), which aims to localize the object with following characteristics: (1) amorphous shape with indistinct boundary; (2) similarity to surroundings; (3) absence in color. Accordingly, it is far more challenging to distinguish insubstantial objects in a single static frame and the collaborative representation of spatial and temporal information is crucial. Thus, we construct an IOD-Video dataset comprised of 600 videos (141,017 frames) covering various distances, sizes, visibility, and scenes captured by different spectral ranges. In addition, we develop a spatio-temporal aggregation framework for IOD, in which different backbones are deployed and a spatio-temporal aggregation loss (STAloss) is elaborately designed to leverage the consistency along the time axis. Experiments conducted on IOD-Video dataset demonstrate that spatio-temporal aggregation can significantly improve the performance of IOD. We hope our work will attract further researches into this valuable yet challenging task. The code will be available at: \url{https://github.com/CalayZhou/IOD-Video}.
\end{abstract}

\section{Introduction}
\label{sec:intro}
Recently, the emergence of deep learning based approaches \cite{girshick2015region, girshick2015fast, ren2015faster} has witnessed significant advancements of object detection. Nevertheless, they still face intractable problems on some insubstantial objects captured by multispectral cameras \cite{hagen2013video} under specific wavelength, e.g. smoke, steam and gas leak. Due to frequent occurrences of smoke poisoning, fire accident, toxic gas leakage and explosion, it is urgent and crucial to realize real-time intelligent monitoring as well as early warning for insubstantial objects. This research topic is fresh and challenging, as insubstantial objects are quite different from conventional objects from several aspects.


\begin{figure}[t]
	\centering
	\includegraphics[width=1.0\linewidth]{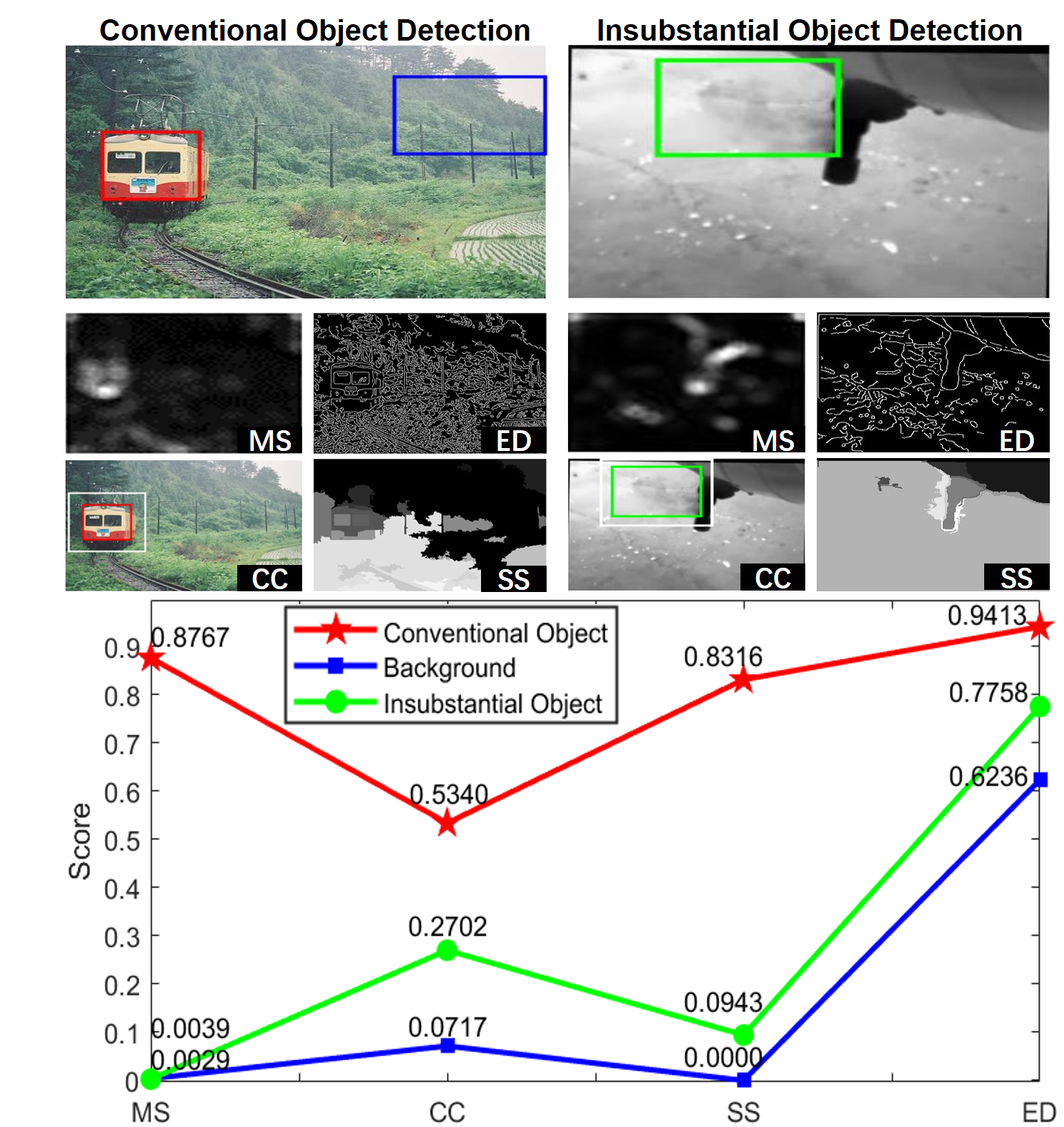}
	\vspace{-1.5em}
	\caption{Comparisons between the conventional object and insubstantial object. The insubstantial object is more  similar to the background  under MS, CC, SS and ED image cues.}
	\label{fig:IODFig}
\end{figure}

 Classical paper \textit{what is an object} \cite{alexe2010object} defined a measure of objectness generic over classes which regards objects as standalone things with a well-defined boundary and center. It is considered that any object has at least one of three distinctive characteristics: (1) a well-defined closed boundary in space; (2) a different appearance from their surroundings; (3) sometimes it is unique within the image and stands out as salient. Based on the above observations, Alexe et al. \cite{alexe2010object} proposed four image cues to distinguish objects: Multi-scale Saliency (MS),  Color Contrast (CC), Edge Density (ED) and Superpixels Straddling (SS). MS \cite{hou2007saliency} indicates that an object is the salient region with a unique appearance; CC reflects the color dissimilarity between the foreground and background; ED measures the average edge magnitude as closed boundary characteristics; SS segments an image into small regions with uniform color or texture. As illustrated in Fig.~\ref{fig:IODFig}, The foreground (train, in red box) in the left achieves relatively high scores for the four measurements, while the background (forest, in blue box) is the opposite. Considering the gas leak (in green box) in the right, it is deprived of trichromatic information owing to the monochromaticity of infrared images. Moreover, the shape of gas leak changes over time and there is no fixed and clear boundary. In terms of above image cues, the gas leak is more similar to the background rather than the foreground. Accordingly, mature algorithms for conventional object detection may fail in this special case, so that specific dataset and algorithm for insubstantial object are urgently needed. 
 
 To facilitate the study on this challenging problem, we collect a video-level insubstantial object detection (IOD-Video) dataset via multispectral camera under various scenes, including the smoke from chimney, hot steam, gas leak, etc. The object characteristics in IOD-Video dataset are summarized as follows: (1) indistinct boundary and amorphous shape; (2) the similarity to the background surroundings; (3) absence of color information and saliency. Consequently, the collaborative representation of spatial and temporal features is crucial under the limitation of spatial information within a static frame. 
 
 Additionally, we develop a spatio-temporal aggregation framework for IOD task. Unlike conventional object detection, IOD should distinguish visual temporal variation more than static semantic appearance. Traditional video background subtraction methods such as Gaussian Mixture Model (GMM) \cite{stauffer1999adaptive} and Visual Background Extractor (Vibe) \cite{barnich2010vibe} need static background, while the optical flow methods \cite{brox2004high} require the moving target to have prominent feature points. They are not applicable to capture the time-variant pattern of insubstantial objects. Deep learning based methods for exploiting spatio-temporal features mainly focus on the classification field, e.g. action recognition \cite{zhu2020comprehensive, chen2021deep}. Moreover, video object detection \cite{jiao2021new} dedicated to propagating the plentiful information from key frame features to non-key frame features and exploiting temporal modeling. The majority paradigm of above approaches first extracts features of a single frame through 2D convolutional network (2D-CNN), and then mines the temporal relation based on the extracted features. In this case, action detection \cite{kalogeiton2017action} is close to IOD, but it is dependent on sufficient spatial information in a single frame and is unsatisfactory for insubstantial object. In this paper, we explore a spatio-temporal aggregation framework from two aspects: First, representative spatio-temporal backbones of action recognition are introduced into our framework to evaluate their accuracy on IOD task. Second, spatio-temporal aggregation loss (STAloss) is designed to impose constraints in the three-dimensional space, as traditional 2D detection losses only concentrate on the static frame and do not take the concordance along time dimension into account. Experimental results reveal that temporal shift models have the best video-level detection performance which may preserve the feature-level integrity of spatial dimension, and STAloss can improve the performance significantly.
 
Our contributions are summarized as follows:

$\bullet$ We propose an IOD-Video dataset for insubstantial object detection to promote research on this challenging task.

$\bullet$ We develop a spatio-temporal aggregation framework in which the video-level detection capability of representative action recognition backbones can be fairly evaluated.

$\bullet$ Based on the temporal shift backbone which achieves best performance, the STAloss is specifically designed to leverage the temporal consistency for further improvement.

\section{Related Work}
\noindent\textbf{Feature Extraction in Action Recognition.} Two-stream networks, 3D-CNNs and compute-effective 2D-CNNs are explored to extract spatio-temporal information in action recognition. The classic paradigm of two-stream network utilized extra modalities that reflect temporal motion information as the second input pathway, e.g. optical flow \cite{wang2016temporal, cheron2015p}. 3D-CNNs are applied to the extraction of spatio-temporal information without pre-computing the input stream which explicitly represents temporal information \cite{tran2015learning, carreira2017quo, hara2018can, diba2018spatio}. Nevertheless, extracting spatio-temporal representation by 3D convolution kernel is of high computational cost, several works \cite{qiu2017learning, tran2018closer} explored the idea of 3D factorization to reduce the complexity. Another straightforward and effective method is to extract frame-level features with 2D-CNNs and then model temporal correlation \cite{lin2019tsm, fan2019more, wang2021tdn, li2020tea, liu2020teinet, zhang2020pan, crasto2019mars, stroud2020d3d}. TSM \cite{lin2019tsm}, TAM \cite{fan2019more} extended the temporal shift operation to video understanding. TIN \cite{shao2020temporal} further fused temporal dependency via interlacing spatial representation. TEA \cite{li2020tea} employed motion excitation and multiple temporal aggregation to enhance the motion pattern, which is similar to the channel-wise spatio-temporal module proposed by STM \cite{jiang2019stm}. TDN \cite{wang2021tdn} adopted the low-cost difference operation to establish multi-scale motion temporal representation by utilizing two temporal difference modules. The study of spatio-temporal collaborative representation mostly focuses on the classification task, there are few attempts in the video-level detection task.

\noindent\textbf{Video Object Detection.} Feature degradation (e.g., motion blur, occlusion, and defocus) is the primary challenge of video object detection (VOD). Early box-level VOD methods tackle this problem in a post-processing way which links bounding boxes predicted by still frames \cite{kang2017t, han2016seq, feichtenhofer2017detect, chen2018optimizing}. Feature-level VOD methods aggregate temporal contexts to improve the feature representation. STSN \cite{bertasius2018object} predicted sampling locations directly and adopted deformable convolutions across space and time to leverage temporal information. STMN \cite{xiao2018video} built the spatio-temporal memory to capture the long-term appearance and motion dynamics. Methods in  \cite{zhu2017flow, wang2018fully, zhu2018towards, jin2022feature,zhu2017deep} aligned and warped adjacent features under the guidance of optical-flow. Besides optical flow, some approaches  \cite{wu2019sequence, deng2019relation, luo2019object, shvets2019leveraging, geng2020object, yao2020video, han2020mining, shvets2019leveraging, koh2021joint, deng2019object} enhanced the object-level feature by exploring semantic and spatio-temporal correspondence among the region proposals. PSLA \cite{guo2019progressive} applied self-attention mechanisms in the temporal-spatial domain without relying on extra optical flow. LSTS \cite{jiang2020learning} adaptively learned the offsets of sampling locations across frames. MEGA \cite{chen2020memory} and LSFA \cite{wang2021real} enhanced global-local aggregation by modeling longer-term and short-term dependency. TF-Blender \cite{cui2021tf} depicted the temporal feature relations and blended valuable neighboring features. TransVOD \cite{he2021end} first introduced temporal Transformer in VOD task to aggregate both the spatial object queries and the feature memories of each frame. Due to the redundant computational cost caused by applying still object detectors to each individual frame, the study of previous VOD methods focuses on propagating the rich information from key frame features to non-key frame features. 

\noindent\textbf{Spatio-temporal Action Detection.} Different from VOD, action detection task pays attention to identifying and localizing human actions in the video. The frame-level detectors \cite{wang2016actionness, peng2016multi}
generate final action tubes by linking frame level detection results. To take full advantage of temporal information, some clip-level detectors have been proposed \cite{kalogeiton2017action, li2018recurrent, gu2018ava, sun2018actor, yang2019step, wu2019long, li2021time, sarmiento2021spatio, liu2021acdnet, pan2021actor}. ACT \cite{kalogeiton2017action} handled a short sequence of frames and output action tubes by regressing from anchor cuboids.  ACRN \cite{sun2018actor} computed actor-scene pair relation information for action classification. Context-Aware RCNN \cite{wu2020context} rethought the importance of resolution in actor-centric and MOC \cite{li2020actions} treated an action instance as a trajectory of moving points. Sarmiento et al. \cite{sarmiento2021spatio} introduced two cross attention blocks to effectively model the spatial relations and capture short range temporal interactions. In recent years, some approaches attempt to recognize action based on 3D convolution features \cite{hou2017tube, yang2019step, pan2021actor,li2021time, liu2021acdnet}. ACDnet \cite{liu2021acdnet} intelligently exploited the temporal coherence between successive video frames to approximate their CNN features rather than naively extracting them. TFNet \cite{li2021time} applied attention mechanism to fuse the temporal features extracted by 3D-CNN and the frequency features extracted by 2D-CNN. ACAR-Net \cite{pan2021actor} built upon a novel high-order relation reasoning operator and an actor-context feature bank to enable indirect relation reasoning. STEP \cite{yang2019step} proposed a progressive approach to obtained high-quality proposals gradually by incorporating more related temporal context. Nevertheless, they lack analysis of how different 3D-CNN architecture designs influence the final performance.

In conclusion, previous VOD and action detection methods mostly rely on 2D-CNN to propose bounding boxes, this paradigm may be unsatisfactory on the IOD task which needs collaborative extraction of spatio-temporal features. 



\section{IOD-Video Dataset}
\subsection{Data Collection and Annotation}
\noindent\textbf{Data Collection.} As most Volatile Organic Compound (VOC) gases do not appear in the visible spectrum and hence they can not be seen by human eyes or traditional RGB cameras. 
The characteristic absorption peaks of many VOC gases are concentrated in the mid infrared spectrum, which is considered as the fingerprint region. So the IOD-Video dataset is captured in a restrained portion of the infrared (IR) domain range in $3\sim5\mu m$ and  $8\sim12\mu m$. Specifically, part of the IOD-Video samples are collected from active deflation experiments of chemical gases by infrared spectral imaging or portable devices. The rest are the real-world samples captured by the alerting or monitoring system deployed in different petrochemical factories, which display various insubstantial objects, including the smoke emission, the water vapor and VOC gas leakage (e.g. olefins, alkanes, carbon monoxide). After continuous collection for nearly three years, we obtained thousands of videos which are manually cleaned up as follows: (1)  The original videos are  cropped into multiple short clips around 10 seconds, each frame in clips is ensured to contain the insubstantial object. (2) At most two representative clips are reserved for the same scene to ensure the diversity of dataset. (3) Eliminate clips that cannot be identified by human eyes or have severe imaging noise. We finally get 600 video samples, amount to a total of 141,017 frames. 

\noindent\textbf{Dataset annotation.} IOD-Video is carefully labeled by three experienced experts with bounding boxes, which are intuitive and practicable. A specifically developed tool is used to improve the annotation quality, by providing the pseudo color, motion information extracted by background modeling and historical annotated frames across two seconds, since longer video-level sequences can significantly improve the ability of human eyes to identify the insubstantial object. We provide frame-level bounding annotations which are double-checked to avoid inconsistency by following rules: (1) Annotations are temporally continuous without sudden change. (2) Bounding boxes tightens the object boundary well by human’s subjective perception. (3) Bounding boxes reacts immediately when diffusion direction varies. The samples are labeled every five frames and middle frames are interpolated due to the slight difference between adjacent frames. All captured insubstantial objects are integrated into one category as foreground.

\begin{figure}[ht]
	\begin{center}
		\includegraphics[width=1.025\linewidth]{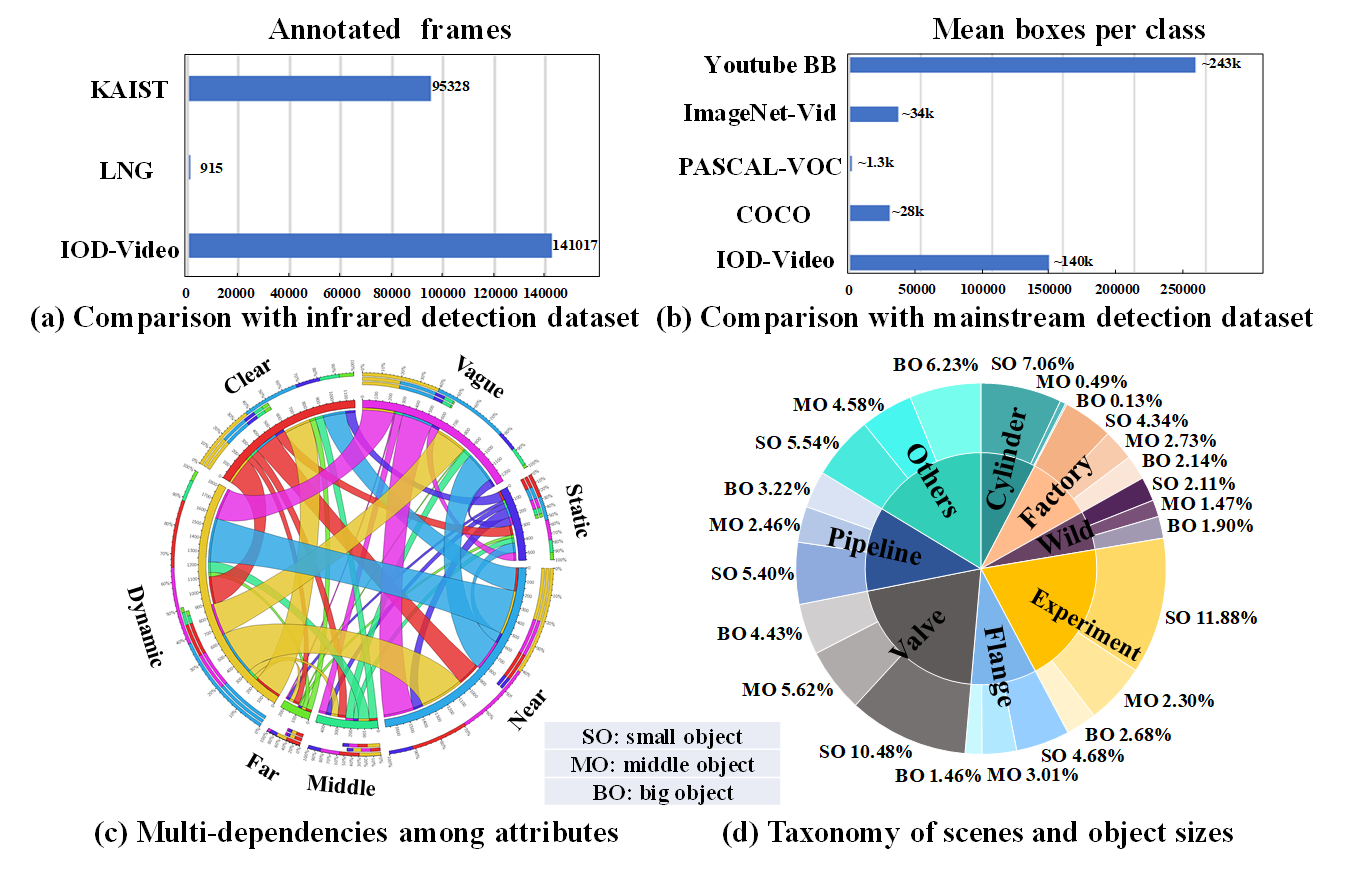}
	\end{center}
	\vspace{-1.7em}
	\caption{ Statistics of the IOD-Video dataset and comparisons with other infrared and main stream detection datasets.}
	\label{fig:dataset statistics}
			\vspace{-0.5em}
\end{figure}

\noindent\textbf{Dataset Statistics.} Although insubstantial objects of IOD-Video dataset captured in mid infrared band are extremely difficult to collect, the annotated frames of IOD-Video exceed the quantity of KAIST \cite{hwang2015multispectral} multispectral pedestrian detection and LNG \cite{bin2021tensor} gas leakage dataset, as shown in Fig.~\ref{fig:dataset statistics} (a). In terms of average boxes per class, IOD-Video is no less than the RGB detection datasets such like COCO \cite{lin2014microsoft}, PASCAL-VOC 
\cite{everingham2010pascal}, ImageNet-Vid \cite{russakovsky2015imagenet}, as shown in Fig.~\ref{fig:dataset statistics} (b). Fig.~\ref{fig:dataset statistics} (c) demonstrates multi-dependencies among IOD-Video attributes, which achieves good diversity by providing various distances (0$\sim$100m), sizes, visibility, and scenes captured by different spectral ranges. The lager width of a link between two super-classes indicates a higher probability.  For example, the dynamic background samples in the near distance occupy a larger proportion than the static background samples in the far distance. IOD-Video dataset covers a wide range of scenarios (including pipeline, factory, flange, valve, experiments, cylinder, wild and others) and objects with different size (e.g., small, middle and big). The distributions of scene categories and object size are displayed in Fig.~\ref{fig:dataset statistics} (d).

\subsection{Challenges}
\begin{figure}[ht]
	\begin{center}
		\includegraphics[width=1.0\linewidth]{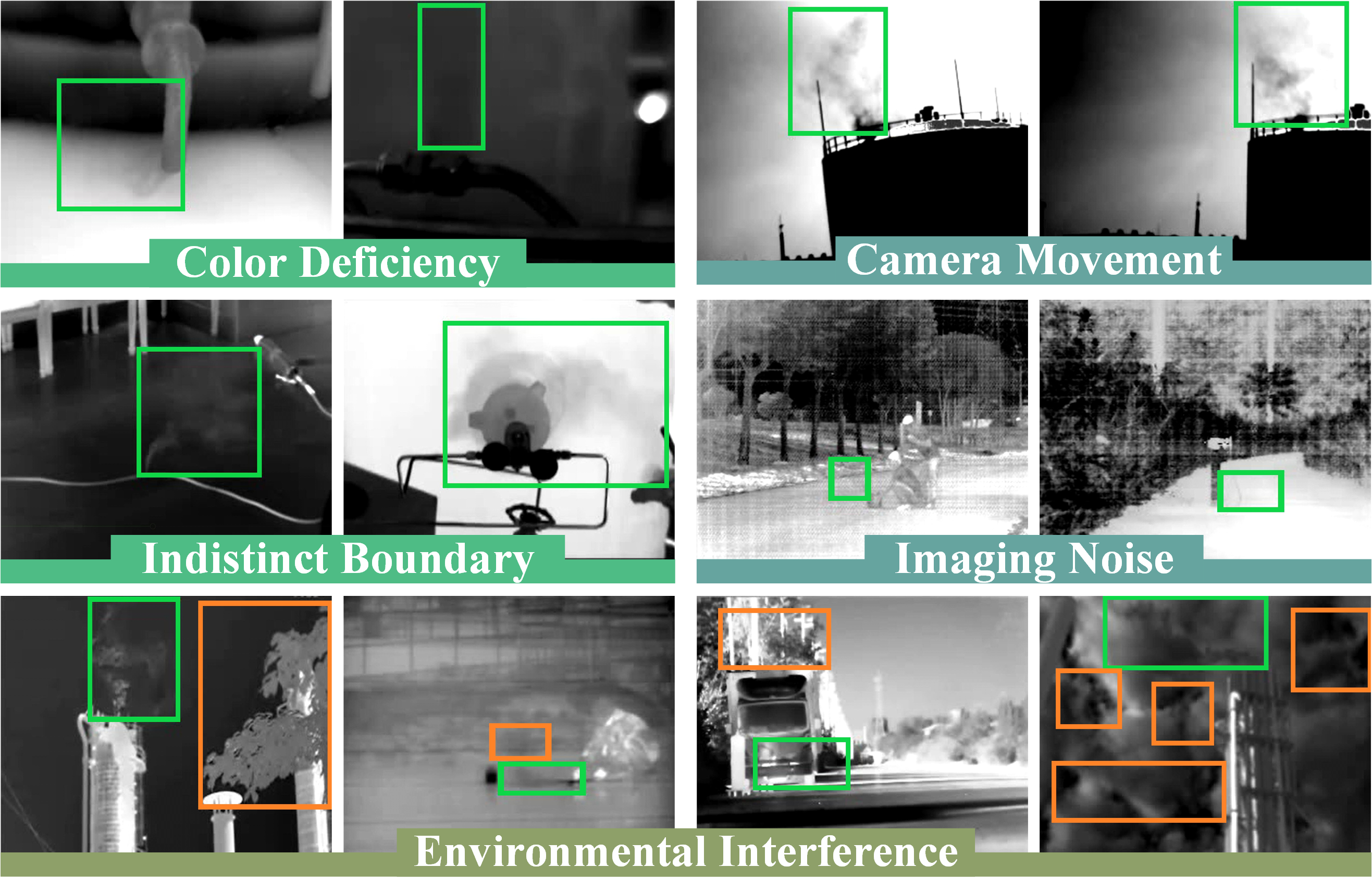}
	\end{center}
	\vspace{-1.5em}
	\caption{Three main challenges: insubstantial characteristics (color absence, indistinct boundary); photography restrictions (camera movement, imaging noise);  environmental interference.}
	\label{fig:challenges}
			\vspace{-0.5em}
\end{figure}

As shown in Fig.~\ref{fig:challenges}, IOD faces multiple challenges caused by its characteristics , photography restrictions and environmental interference. First, the color absence in infrared video samples with indistinct boundary makes it intractable to identify and localize insubstantial object. 
Second, some IOD-video samples are captured by patrol inspection devices and the camera movement makes it difficult to locate the insubstantial object caused by scene switching and camera shaking. Due to temperature drift caused by ambient temperature and infrared radiation absorption, the infrared cameras (e.g., cooled infrared detection arrays and uncooled infrared focal plane arrays suffer from low signal-to-noise ratio for infrared inhomogeneity and various imaging noises especially in the absence of heat source.
Third, insubstantial objects sometimes appear to be invisible when it coincides with the complex background. In addition, the impact of environmental interference will be magnified due to the monochromaticity of infrared images, such as leaves, grasses, dusts blown by the wind.

\begin{figure*}
	\begin{center}
		\includegraphics[width=1.0\linewidth]{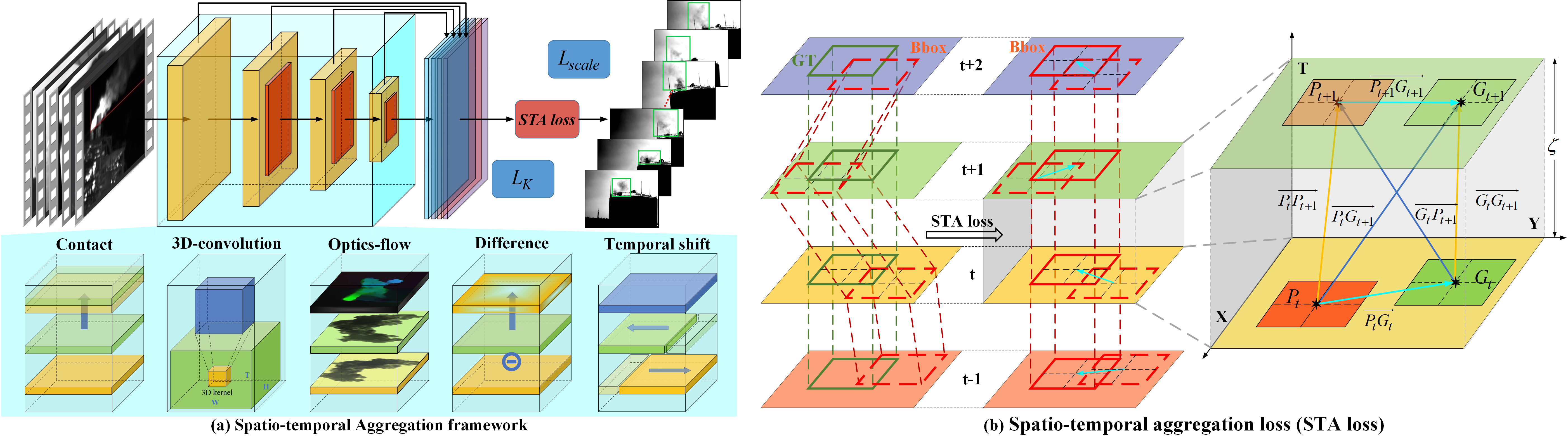}
	\end{center}
	\vspace{-1.5em}
	\caption{(a) Our framework is inserted with five different designed spatio-temporal backbones to evaluate the detection capability of action recognition models. (b) The STAloss acts as an extra constraint to gather the predicted boxes (red dashed boxes) to the true boxes (green solid boxes) along the temporal dimension. The final results adjusted by STAloss are shown as red solid boxes.}
	\label{fig:framework}
		\vspace{-0.5em}
\end{figure*}

\subsection{Evaluation Metrics and Protocols}

The IOD-Video dataset samples are divided into clear and vague sets according to whether annotator can judge the boundary of the object within a single frame subjectively.
We refer the COCO evaluation protocol \cite{lin2014microsoft} and report average precision over all IOU thresholds (AP), AP at IOU thresholds 0.5 (@0.5), 0.75 (@0.75), clear set (@clear) and vague set (@vague) in the frame level. In addition, IOD-Video dataset is randomly divided into three split (train/test split at a ratio of 2:1) and K-Fold cross-validation is adopted to report results averaged over three splits following the common setting \cite{peng2016multi, kalogeiton2017action, li2020actions}.

\section{Spatio-temporal Aggregation Framework}
\subsection{Overview}
Insubstantial objects can hardly be located in a single static frame, as they differ from convectional objects with unclear contours. Thus, IOD needs  spatio-temporal collaborative representation of several adjacent frames which remains unexplored in video-level detection task before. In this section, we design a general video-level detection framework as illustrated in Fig.~\ref{fig:framework}, where representative action recognition models can be adopted as the spatio-temporal backbone. The anchor-free model CenterNet \cite{zhou2019objects} is employed as the basic architecture to make the whole pipeline as simple as possible. All spatio-temporal backbones are built on ResNet-50 \cite{he2016deep}, hence the detection performance of different action recognition models on IOD can be fairly compared with our framework. To be specific, multi-scale feature maps from ResNet-50 $stage$ $2,3,4,5$ are up-sampling with the deconvolution layer into the same resolution, $T$ images with resolution of $W\times H$ are fed into the spatio-temporal backbone to generate a feature volume $\mathcal{F}^{T\times\frac{W}{R} \times \frac{H}{R}\times 64}$, $R$ is spatial down-sampling ratio. In order to make full use of annotations, the original loss of base detector CenterNet is calculated based on the output feature volume $\mathcal{F}$ for each input frame.
Since IOD needs spatio-temporal representation in feature extraction stage, naturally, the loss function design should also make corresponding changes. Then aiming to leverage the temporal consistency, we design the STAloss that will be presented in section~\ref{sec:staloss}. Next, the technical details of these spatio-temporal backbones will be presented.

\subsection{Spatio-temporal Backbone}
We choose state-of-the-art approaches \cite{wang2021tdn, li2020tea, fan2019more, lin2019tsm, kwon2020motionsqueeze, shao2020temporal, carreira2017quo, xie2018rethinking} on Something-Something (Sth-Sth) dataset \cite{goyal2017something} as spatio-temporal backbones. Analogous to IOD dataset, the static background in Sth-Sth dataset contributes little to the final prediction \cite{zhu2020comprehensive}, and strong motion reasoning is required to capture the long-term temporal structure.  

\noindent\textbf{Concat.} Direct concatenation of input $T$ frame features extracted by 2D-CNN is the simplest way to model temporal information. We follow the concatenation implementation of previous MOC \cite{li2020actions} methods. 

\noindent\textbf{3D-CNN.} 3D-CNN \cite{carreira2017quo, xie2018rethinking} jointly learns spatial and temporal features by extending standard 2D spatial convolutional networks to the temporal dimension. Nevertheless, it brings higher computational cost and lacks specific consideration in the temporal information.

\noindent\textbf{Flow-based.} Many action recognition methods utilize the pre-computed dense optical flow as explicit motion representation. We argue it may fail on insubstantial object for the lack of outstanding feature points, and therefore Motion Squeeze Network (MSNet) \cite{kwon2020motionsqueeze} is adopted to establish flow correspondences implicitly.

\noindent\textbf{Temporal Difference.} As an approximate motion representation, temporal difference explicitly computes motion information and captures the distinctive properties of the adjacent frames. Numerous methods \cite{wang2021tdn, wang2016temporal, zhao2018recognize} have shown effectiveness in action recognition by utilizing the temporal difference operator into the network design.  

\noindent\textbf{Temporal Shift.} The temporal dynamics are embedded into spatial representations by shifting the channels along the temporal dimension, which is a simple yet effective design with strong spatio-temporal modeling ability.

%

\begin{table*}
	\begin{center}
		\renewcommand{\arraystretch}{1.2}
		\setlength{\tabcolsep}{2.3mm}{
			\begin{tabular}{c|c|c|c|ccccc}
				\hline
				\multirow{2}{*}{\begin{tabular}[c]{@{}c@{}}\\ \end{tabular}}                & \multicolumn{1}{c|}{\multirow{2}{*}{Method}} & \multicolumn{1}{c|}{\multirow{2}{*}{Base Backbone}} & \multicolumn{1}{c|}{\multirow{2}{*}{\begin{tabular}[c]{@{}c@{}}Pretrained\\ Model\end{tabular}}} & \multicolumn{5}{c}{Frame AP (\%)}                                                                                                            \\ \cline{5-9} 
				& \multicolumn{1}{c|}{}                        & \multicolumn{1}{c|}{}                          & \multicolumn{1}{c|}{}                                                                          & \multicolumn{1}{l}{@0.5} & \multicolumn{1}{l}{@0.75} & \multicolumn{1}{l}{@clear} & \multicolumn{1}{l}{@vague} & \multicolumn{1}{l}{0.5:0.95} \\ \hline
				\multirow{3}{*}{\begin{tabular}[c]{@{}c@{}}Frame-based\\ Detector\end{tabular}}            & Faster RCNN \cite{ren2015faster}                                  & ResNet-50                                      & ImageNet                                                                                       & 33.49                    & 6.76                      & 16.52                      & 8.82                      & 12.31                        \\
				& SSD \cite{liu2016ssd}                                        & ResNet-50                                      & ImageNet                                                                                       & 30.21                    & 3.95                     & 12.78                      & 7.82                       & 9.99                        \\
				& CenterNet \cite{zhou2019objects}                                  & ResNet-50                                      & ImageNet                                                                                       & 24.80                    & 4.32                      & 11.21                      & 6.65                       & 8.50                        \\ \hline
				\multirow{3}{*}{\begin{tabular}[c]{@{}c@{}}Video-based\\Detector \\ \end{tabular}} & CRCNN \cite{wu2020context}                                        & ResNet101\&FPN                                 & ImageNet                                                                                       & 36.15                    & 7.46                      & 18.84                      & 9.14                      & 13.52                        \\
				& MOC \cite{li2020actions}                                          & DLA-34                                         & COCO                                                                                           & 36.81                    & \textbf{8.94}                     & 19.96                      & 9.62                      & 14.29                        \\
				& MOC + Flow \cite{li2020actions}                                 & DLA-34                                         & COCO                                                                                           & 34.50                    & 7.28                      & 18.18                      & 8.49                      & 12.75                        \\ \hline
				Ours & TEA \cite{li2020tea}                                         & ResNet-50                                     & ImageNet                               & \textbf{42.19}                   & 8.66                     & \textbf{22.97}                     & \textbf{9.95}                & \textbf{15.69}                    \\ 
 \hline
		\end{tabular}}
	\end{center}
		\vspace{-1.5em}
	\caption{Comparisons with previous frame-based detectors and video-based detectors. Our basic spatio-temporal aggregation framework simply replaces the backbone of CenterNet with TEA \cite{li2020tea} without any other complex design.}
	\label{STOA}
			\vspace{-0.5em}
\end{table*}
\subsection{Spatio-temporal Aggregation Loss}
\label{sec:staloss}
After getting predicted boxes for each input frame, they may have a good regression within single static frames, but the consistency of predicted results along the time dimension is not subject to constraints. For instance, all four predicted boxes in Fig.~\ref{fig:framework} (b) has the same Intersection-over-Union (IoU) with the ground truth (GT) in terms of spatial dimension. Nevertheless, they are staggered in the time axis. The STAloss is able to pull the predicted boxes to the true boxes across multi frames.

In view of above observations, STAloss (i.e., $\mathnormal{L}_{STA}$) is proposed to impose constraints in the three-dimensional space. $\mathnormal{L}_{STA}$ is consisted of $\mathnormal{L}_{STA\cos\theta}$ and  $\mathnormal{L}_{STA\sin\beta}$, they can be optimized in a collaborative manner. Let $P_{t}^{pre}(x_{pc}^{t}, y_{pc}^{t}, t)$ be the predicted box center of the $t^{th}$ frame and $G_{t}(x_{gc}^{t}, y_{gc}^{t}, t)$ be the corresponding GT center, an extra STA branch is established based on the output feature volume $\mathcal{F}$ to predict the offset  $( \Delta^{t}_{x}, \Delta^{t}_{y}, t)$, which adjusts the box center of each input frame to a proper location $P_{t}(x_{pc}^{t} + \Delta^{t}_{x}, y_{pc}^{t} + \Delta^{t}_{y},t)$ and the adjusted box of next frame is $P_{t+1}(x_{pc}^{t+1} + \Delta^{t+1}_{x}, y_{pc}^{t+1} + \Delta^{t+1}_{y},t+\zeta)$ .  $\zeta$ is the only hyperparameter in $\mathnormal{L}_{STA}$ which represents temporal interval length between adjacent frames. $L_{STA\cos \theta }^{cross}$ represents the dot multiplication of cross vectors $\overrightarrow {{G_t}{P_{t + 1}}}$ and $\overrightarrow{{P_t}{G_{t + 1}}}$, which will pull the prediction boxes towards GT boxes when they are far away. The $L_{STA\cos \theta }^{self}$ restricts the vector direction of the center line of adjacent prediction boxes $\overrightarrow {{P_t}{P_{t + 1}}}$ to be consistent with the GT boxes $\overrightarrow {{G_t}{G_{t + 1}}}$.  $\theta$ reflects the angle between the cross vectors $\overrightarrow{{P_t}{G_{t + 1}}}$,  $\overrightarrow {{G_t}{P_{t + 1}}}$ and the angle between self vectors $\overrightarrow {{P_t}{P_{t + 1}}}$, $\overrightarrow {{G_t}{G_{t + 1}}}$. The angle $\theta$ is expected to approach $0^{\circ}$ during training. The offset is optimized by the following items:
\begin{equation}
\left\{ {\begin{array}{*{20}{c}}
	{ L_{STA\cos \theta }^{cross} = \frac{1}{{T-1}}\sum\limits_{t=1}^{T-1} { {\frac{{\overrightarrow {{G_t}{P_{t + 1}}}  \cdot \overrightarrow {{P_t}{G_{t + 1}}} }}{{\left| {\overrightarrow {{G_t}{P_{t + 1}}} } \right|\left| {\overrightarrow {{P_t}{G_{t + 1}}} } \right|}}} } } \hfill  \vspace{1ex}  \\
	{ L_{STA\cos \theta }^{self} = \frac{1}{{T-1}}\sum\limits_{t=1}^{T-1} { {\frac{{\overrightarrow {{P_t}{P_{t + 1}}}  \cdot \overrightarrow {{G_t}{G_{t + 1}}} }}{{\left| {\overrightarrow {{P_t}{P_{t + 1}}} } \right|\left| {\overrightarrow {{G_t}{G_{t + 1}}} } \right|}}} } } \hfill   \vspace{1ex} \\
	{ L_{STA\sin \beta }^{pre} = \frac{1}{{T-1}}\sum\limits_{t=1}^{T-1} {{\frac{{\left| {\overrightarrow {{P_t}{G_t}} } \right|}}{{\left| {\overrightarrow {{P_t}{G_{t + 1}}} } \right|}}} } } \hfill  \vspace{1ex} \\
	{ L_{STA\sin \beta }^{next} = \frac{1}{{T-1}}\sum\limits_{t=1}^{T-1} { {\frac{{\left| {\overrightarrow {{P_{t + 1}}{G_{t + 1}}} } \right|}}{{\left| {\overrightarrow {{P_{t + 1}{G_t}}} } \right|}}} } } \hfill 
	\end{array}} \right.
\end{equation}

 \noindent while $\cos\theta$  trends to be a smooth curve when  $\theta \rightarrow 0^{\circ}$, this may cause difficulty to further converge. To address this issue, we introduce a $L_{STA\sin \beta}$ term which is the division of ${\left| {\overrightarrow {{P_{t }}{G_{t }}} } \right|}$ and ${\left| {\overrightarrow {{P_{t}}{G_{t + 1}}} } \right|}$ inspired by Distance-IoU loss \cite{zheng2020distance}. The vector $\overrightarrow {{G_t}{G_{t + 1}}}$ is almost perpendicular to $XY$ plane due to the tiny changes between adjacent frames in most cases, so $L_{STA\sin \beta}$ term can be approximate  equal to $\sin \beta$ where $\beta$ is the angle between the vector ${\overrightarrow {{P_{t}}{G_{t+1}}}}$ and ${\overrightarrow {{G_{t}}{G_{t+1}}}}$. When the predicted boxes and  GT boxes become closer during the later training process, the $L_{STA\sin \beta }$ term will locate the center more accurately in term of steeper gradient when angle $\beta \rightarrow 0^{\circ}$. The cooperative effect of $L_{STA\cos \theta }$ and $L_{STA\sin \beta }$ leverage the temporal consistency along the time dimension. We average the above terms and final $\mathnormal{L}_{STA}$ is formulated as:

\begin{equation}
\left\{ {\begin{array}{*{20}{c}}
	{{L_{STA\cos \theta }} = \frac{1}{2}L_{STA\cos \theta }^{cross} + \frac{1}{2}L_{STA\cos \theta }^{self}} \hfill  \vspace{1ex} \\
	{{L_{STA\sin \beta }} = \frac{1}{2}L_{STA\sin \beta }^{pre} + \frac{1}{2}L_{STA\sin \beta }^{next}} \hfill  \vspace{1ex} \\
	{{L_{{STA}}} = \lambda\left( {1-{L_{STA\cos \theta }}} \right) + (1-\lambda){L_{STA\sin \beta }}} 
	\end{array}} \right.
\end{equation}

\noindent where $\lambda$ is the hyperparameter between the ${L_{STA\sin \beta }}$ and ${L_{STA\cos \theta }}$, empirically we set $\lambda$ to 0.5. The overall training objective is:
\begin{equation}
L = {L_K} + {\lambda_{size}L_{size}} + {L_{STA}}\
\end{equation}

\noindent where ${L_K}$ and $L_{size}$ is the original classification loss and scale loss of the base detector CenterNet \cite{zhou2019objects}, and ${L_{STA}}$ acts as an extra loss which can be inserted into any other video-level detection pipeline.

\section{Experiments}

\subsection{Implementation Details}

 We apply the same data augmentation with MOC \cite{li2020actions} to the whole training video clips: mirroring, distorting, expanding and cropping. Specifically, in the training we crop a patch with the size of [0.3, 1] of the input image and resize it to 288 $\times$ 288, then each image is randomly distorted and horizontal flipped with a probability of 0.5 to increase the diversity. The spatial down-sampling ratio $R$ is set to 4 and the temporal interval length $\zeta$ is set to 4. The whole network is trained by Adam optimizer with a learning rate of 5e-4 and a batch size of 16 on two nvidia 3090 GPUs, we decrease the learning rate by 0.1$\times$ at the 6th and 8th epochs, and stop at the 12th epoch. For video-based detectors, the number of input frames is set to 8 unless otherwise stated.

\begin{table*}[t]
	\begin{center}
		\renewcommand{\arraystretch}{1.0}
		\setlength{\tabcolsep}{3.8mm}{
			\begin{tabular}{c|c|c|ccccc}
				\hline
				\multirow{2}{*}{}       & \multirow{2}{*}{\begin{tabular}[c]{@{}c@{}}Spatio-temporal\\ Backbone\end{tabular} } & \multicolumn{1}{c|}{\multirow{2}{*}{\begin{tabular}[c]{@{}c@{}}Base\\ Backbone\end{tabular}}} & \multicolumn{5}{c}{Frame AP (\%)}               \\ \cline{4-8} 
				&                          & \multicolumn{1}{c|}{}                          & @0.5    & @0.75   & @clear  & @vague  & 0.5:0.95 \\ \hline
				Concat                          & Concat \cite{li2020actions}                  & ResNet-50                                      & 27.41 & 4.45  & 12.38 & 7.32  & 9.32  \\ \cline{1-1}
				\multirow{2}{*}{3D-Convolution} & S3D \cite{xie2018rethinking}                      & ResNet-50                                      & 35.82 & 6.73  & 17.72 & 8.81  & 12.72  \\
				& I3D \cite{carreira2017quo}                    & ResNet-50                                      & 36.83 & 7.39  & 18.78 & 9.29  & 13.43 \\ \cline{1-1}
				Flow-based                     & MSNet \cite{kwon2020motionsqueeze}                   & ResNet-50                                      & 41.19 & 7.91  & 21.35 & 10.01 & 14.90  \\ \cline{1-1}
				Difference                      & TDN \cite{wang2021tdn}                     & ResNet-50                                      & 41.69 & 8.48  & 21.42 & 10.46 & 15.40  \\ \cline{1-1}
				\multirow{4}{*}{Temporal Shift} 
				& TSM \cite{lin2019tsm}                      & ResNet-50                                      & 42.13 & 8.20  & 21.98 & 10.28 & 15.38  \\
				& TAM \cite{fan2019more}                    & ResNet-50                                      & 41.95 & 8.53  & 21.49 & 10.61 & 15.50  \\
				& TIN \cite{shao2020temporal}                      & ResNet-50                                      & 42.77 & 8.01 & 22.35 & 10.51 & 15.73  \\
				& TEA \cite{li2020tea} 
				& Res2Net-50                                     & 42.19 & 8.66 & 22.97 & 9.95 & 15.69  \\ \cline{1-8}
				\multirow{2}{*}{+STAloss} & TIN \cite{shao2020temporal}                  & ResNet-50                                      & 43.72& 9.26  & 23.81 & 10.35  & 16.27 \\
				& TEA \cite{li2020tea}                   & Res2Net-50                                      & \textbf{
				45.08} & \textbf{9.50}  & \textbf{24.43} & \textbf{10.91}  & \textbf{16.99} \\ 
				
				\hline
		\end{tabular}}
	\end{center}
		\vspace{-1.5em}
	\caption{The  representative action recognition models are selected from the state-of-the-arts on Sth-Sth dataset \cite{goyal2017something}, and we present the video-level detection performance of different spatio-temporal backbones. STAloss replaces the original offset loss $\mathnormal{L}_{off}$ of CenterNet.}
	\label{stbackbone}
		\vspace{-1.0em}
\end{table*}

  \renewcommand{\arraystretch}{1.2}
\begin{table}[]
	\begin{center}
		\setlength{\tabcolsep}{1.6mm}{
			\begin{tabular}{c|ccccc}
				\hline
				\multirow{2}{*}{\begin{tabular}[c]{@{}c@{}}input \\ frames\end{tabular}} & \multicolumn{5}{c}{Frame AP (\%)}        \\ \cline{2-6} 
				& @0.5 & @0.75 & @clear & @vague & 0.5:0.95 \\ \hline
				2                                                                      & 33.60 & 5.65   & 13.35  & 7.98 & 11.78   \\
				4                                                                        & 40.46 & 8.02  &14.51  & 9.77   & 14.89     \\
				6                                                                        & 44.44 & 9.91 & 24.09  &11.17 & 16.97    \\
				8                                                                         & 45.08 & 9.50 & 24.43  & 10.91   & 16.99     \\ \hline
		\end{tabular}}
		\end {center}
		\vspace{-0.6cm} 
		\caption{Evaluation results with different input frames.}
		\label{temporalrange}
			\vspace{-0.2cm} 
	\end{table}
 
 \subsection{Performance of Classic Detectors}
 
We first analyze detection performances of classic frame-based and video-based detectors on IOD-Video dataset in Tab.~\ref{STOA}. In terms of the frame-based detectors, the video clips are split into frames as training samples. Faster RCNN \cite{ren2015faster} achieves the best results under the condition of the same backbone. On the one hand, compared with SSD \cite{liu2016ssd}, the two-stage architecture design of Faster RCNN proves to be beneficial since there existing a large number of hard negatives in IOD-Video samples. On the other hand, the original deep layer aggregation backbone (DLA-34 \cite{yu2018deep}) is deprived from CenterNet, and this seems to preclude the benefits of hierarchical feature fusion which is important for anchor-free design. Note that all methods in Tab.~\ref{STOA} utilize ImageNet pretrained model except for MOC \cite{li2020actions}, which uses the COCO pretrained model to provide stronger spatial representation ability while reducing sensitivity to dynamic change. In addition, the concatenation design lacks in-depth mining of temporal information. The dense-flow extracted by external off-the-shelf method \cite{brox2004high} usually obtains gains on action recognition tasks but shows minimal improvements on IOD task, due to the shortage of heavily dependent on texture and color features in spatial domain. Our spatio-temporal aggregation framework simply replaces the backbone of CenterNet \cite{zhou2019objects} with TEA \cite{li2020tea}, and it achieves overall better performances which suggest the spatio-temporal aggregation is crucial. 

\begin{figure}[t]
	\centering
	\includegraphics[width=1.0\linewidth]{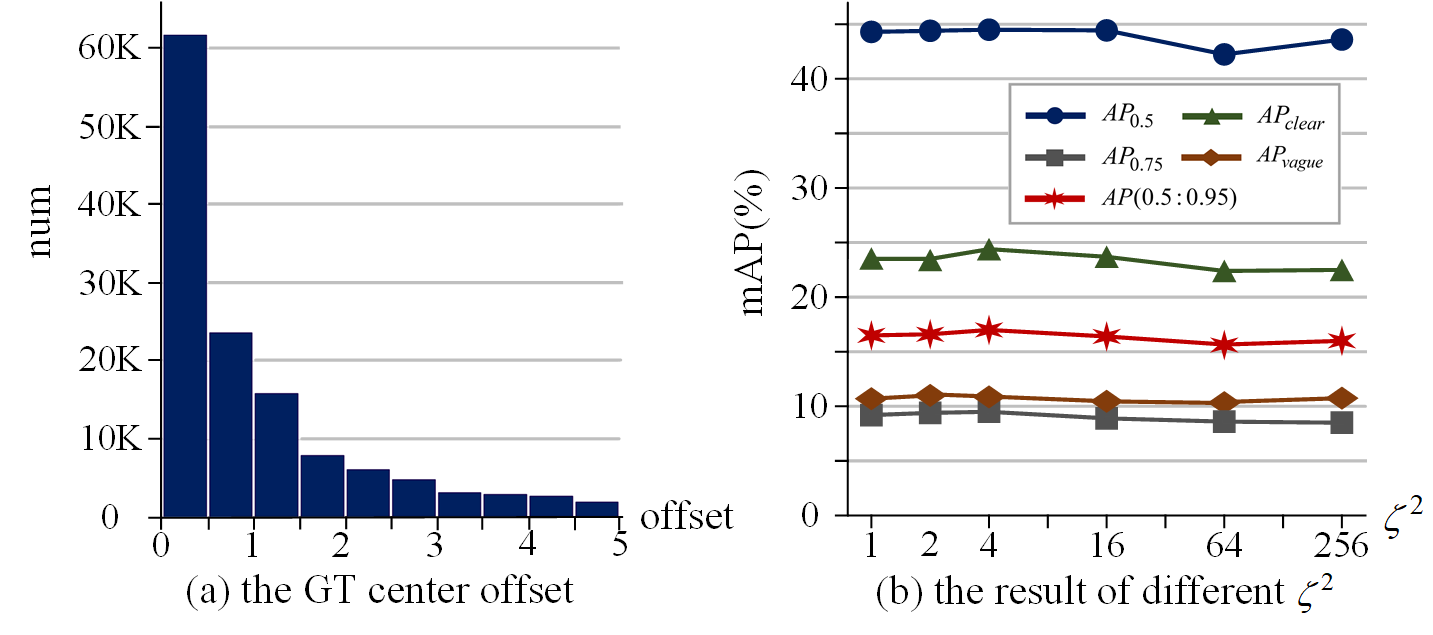}
	\vspace{-1.8em}
	\caption{(a) The statistical GT center offset of IOD-Video dataset. (b) Analysis of different hyperparameter $\zeta^{2}$ value settings. }
	\label{parameter}
			\vspace{-0.5em}
\end{figure}

\begin{figure*}[t]
	\begin{center}
		\includegraphics[width=1.0\linewidth]{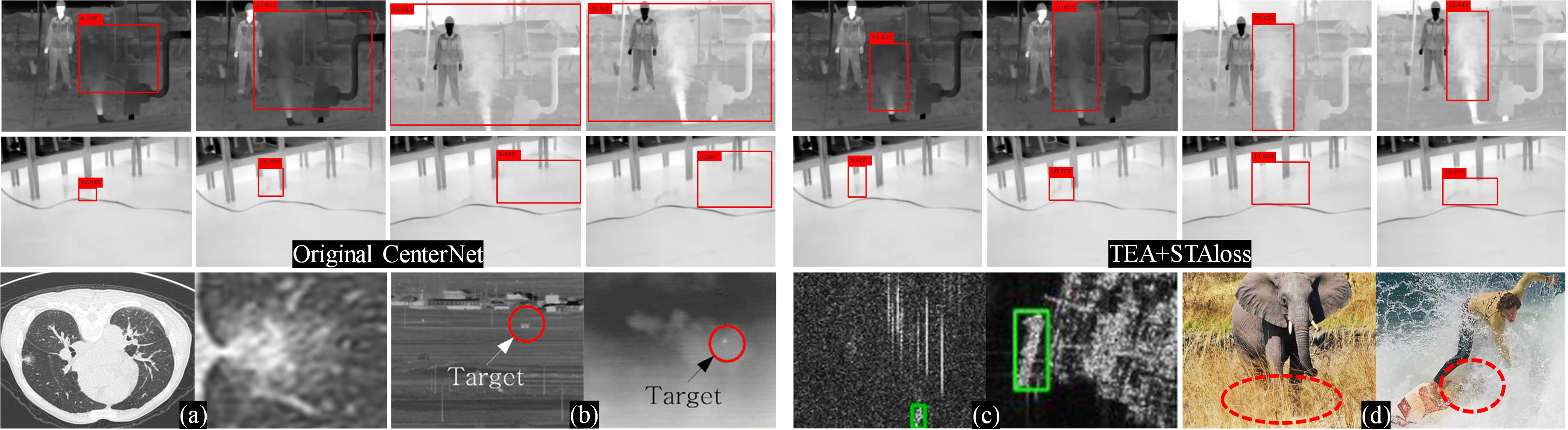}
	\end{center}
	\vspace{-1.5em}
	\caption{Detection results of CenterNet and our baseline method (TEA+STAloss) for clear and vague samples. Potential applications: (a) Medical diagnosis (b) Infrared dim small object detection (c) Synthetic Aperture Radar (SAR) detection (d) Partially-occluded targets. }
	\label{fig:application}
		\vspace{-1.0em}
\end{figure*}
\subsection{Spatio-temporal Backbones}
Next we explore the spatio-temporal backbones with different architecture designs. As shown in Tab.~\ref{stbackbone}, although simple concatenation operation \cite{li2020actions} performs well on the spatio-temporal action detection task, it fails on the IOD task due to the significant difference between two tasks within a single static frame. The action detection datasets, UCF101-24 \cite{soomro2012ucf101} and JHMDB\cite{jhuang2013towards}, tend to be scene-focused while our IOD-Video dataset is motion-focused. 3D convolution \cite{carreira2017quo} or spatial and temporal separable 3D convolution \cite{xie2018rethinking} provide a feasible paradigm in action recognition task, but they lack explicit mining of spatio-temporal datacube along the temporal dimension, which causes performance degradation compared with compute-effective 2D-CNNs. A trainable neural module proposed in MSNet \cite{kwon2020motionsqueeze} has a relatively good performance on the IOD task, which establishes correspondences across multi-frames and converts them into motion features (flow-based). The implicit extraction of motion information may alleviate the difficulty of extracting dense optical flow directly on IOD-Video dataset to some extent. TDN \cite{wang2021tdn} presents a two-level temporal modeling framework to generalize the idea of RGB difference for motion modeling. The similar idea can be found in the design of TEA \cite{li2020tea} structure, which indicates that direct subtraction is a simple yet effective way to capture the insubstantial feature. Moreover, the temporal shift based methods are capable to achieve superior performance. Among them, TSM \cite{lin2019tsm} and TAM \cite{fan2019more} are the initial design which achieve the comparable performance. TIN \cite{shao2020temporal} adopts deformable shift module and has the highest accuracy on AP@[0.5:0.95] without STAloss added. TEA is the combination of subtraction and temporal shift in a light-weight configuration. In our opinion, temporal shift models preserve the feature-level integrity of spatial dimension and can be an effective design for strong temporary motion reasoning as well as spatial semantic representation on video-level detection tasks.

\subsection{The Number of Input Frames }
It is believed that the input range plays an important role to capture the temporal information, we evaluate the performance under different input frames sets. Intuitively, the more input frames to be, the higher AP is expected to be obtained. Nevertheless, the input frames 8 shows tiny advantage compared with input frames 6 in in Tab.~\ref{temporalrange}. It is reflected that our baseline method mainly focuses on short-term temporal modeling, the long-term temporal modeling ability with novel insight is needed for further improvement.

\subsection{STAloss}
To further leverage temporal consistency in the loss level, we apply STAloss to the TEA and TIN which perform best among spatio-temporal backbones. The STAloss can bring a substantially better localization accuracy, especially  at AP@0.5 and AP@clear sets. The comprehensive results over five sets indicate that it is a feasible way to impose constraints in spatio-temporal space of loss function design for IOD task. Since training a video-level detector with the STAloss as auxiliary objective only involves one hyperparameter $\zeta$ in the internal structure of loss function, we conduct several experiments to investigate the robustness of hyperparameter $\zeta$, which is used to adjust the temporal interval length. We make a statistic of the offset between the center of adjacent GT boxes on IOD-Video dataset. The GT offset of a single sample and hyperparameter $\zeta$ are combined to form the vector ${\overrightarrow {{G_{t}}{G_{t+1}}}}$. Fig.~\ref{parameter} (a) shows a sharp downward curve which illustrates the slight change for most cases along the time axis. Under this condition, different values of $\zeta^{2}$ in {1, 2, 4, 16, 64, 256} are set and we observe our baseline is relatively insensitive to the variations of  $\zeta^{2}$ from 1 to 256 in Fig.~\ref{parameter} (b). Combine the statistical curve of GT center offset and $\zeta^{2}$ settings into consideration, $\zeta$ should be larger than most offset statistics empirically for the convergence at the early stage of training. When the $\zeta$ is set too large, the $L_{STA\cos \theta }$ will always be 1 and $L_{STA\sin \beta }$ tend to be 0, which brings difficulty for STAloss to be optimized. Overall, the only hyperparameter $\zeta$ is robust within the appropriate range and the proposed STAloss can be nearly regarded as hyperparameter-free.

\section{Conclusion}

In this work we make attempts on a rarely explored task named insubstantial object detection which is completely different from previous detection tasks. Insubstantial objects have indistinct boundary and amorphous shape, and may appear to be invisible in the background due to the lack of color information. In consideration of this, a feasible way is exploiting spatio-temporal features to compensate the feature absence of single static frames. But the feature extraction of previous video-level detection methods mainly rely on 2D-CNN and this paradigm may be unsatisfactory on IOD. So we collect the IOD-Video dataset which consists of 600 videos (141,017 frames) and construct the spatio-temporal aggregation framework from two aspects: First we measure the detection capacity of different action recognition backbones and reveal temporal shift models perform best; Second the STAloss is designed to pull the prediction boxes of each frame along the temporal dimension together. As shown in Fig.~\ref{fig:application}, compared with CenterNet, our baseline method (TEA backbone + STAloss) is robust to light and shade variations and severe deformation with spatio-temporal aggregation introduced. Nevertheless, there are still large room for our baseline method to improve. Additionally, IOD may benefit the study of infrared dim small objection detection \cite{bae2014spatial}, ground-glass nodule diagnosis in medical \cite{mei2021sanet}, Synthetic Aperture Radar (SAR) detection \cite{zhang2020hyperli} and some partially-occluded targets in RGB images \cite{zheng2021localization}, which also have the similar characteristic of indistinct boundary. We consider to extend the IOD-Video dataset to multispectral bands in the future, which will make it possible to distinguish the specific substance of detected objects. The proposed IOD-Video dataset and baseline approach are expected to drive progress on the study of challenging IOD task.

\noindent\textbf{Acknowledgment.} This research was supported by National Natural Science Foundation of China under Grant (62025108), we thank \emph{ZHIPUTECH} for the multispectral camera supports and data collection.


{\small
\bibliographystyle{ieee_fullname}
\bibliography{egbib}

\begin{thebibliography}{10}\itemsep=-1pt

\bibitem{alexe2010object}
Bogdan Alexe, Thomas Deselaers, and Vittorio Ferrari.
\newblock What is an object?
\newblock In {\em 2010 IEEE computer society conference on computer vision and
  pattern recognition}, pages 73--80. IEEE, 2010.

\bibitem{bae2014spatial}
Tae-Wuk Bae.
\newblock Spatial and temporal bilateral filter for infrared small target
  enhancement.
\newblock {\em Infrared Physics \& Technology}, 63:42--53, 2014.

\bibitem{barnich2010vibe}
Olivier Barnich and Marc Van~Droogenbroeck.
\newblock Vibe: A universal background subtraction algorithm for video
  sequences.
\newblock {\em IEEE Transactions on Image processing}, 20(6):1709--1724, 2010.

\bibitem{bertasius2018object}
Gedas Bertasius, Lorenzo Torresani, and Jianbo Shi.
\newblock Object detection in video with spatiotemporal sampling networks.
\newblock In {\em Proceedings of the European Conference on Computer Vision
  (ECCV)}, pages 331--346, 2018.

\bibitem{bin2021tensor}
Junchi Bin, Choudhury~A Rahman, Shane Rogers, and Zheng Liu.
\newblock Tensor-based approach for liquefied natural gas leakage detection
  from surveillance thermal cameras: A feasibility study in rural areas.
\newblock {\em IEEE Transactions on Industrial Informatics}, 17(12):8122--8130,
  2021.

\bibitem{brox2004high}
Thomas Brox, Andr{\'e}s Bruhn, Nils Papenberg, and Joachim Weickert.
\newblock High accuracy optical flow estimation based on a theory for warping.
\newblock In {\em European conference on computer vision}, pages 25--36.
  Springer, 2004.

\bibitem{carreira2017quo}
Joao Carreira and Andrew Zisserman.
\newblock Quo vadis, action recognition? a new model and the kinetics dataset.
\newblock In {\em proceedings of the IEEE Conference on Computer Vision and
  Pattern Recognition}, pages 6299--6308, 2017.

\bibitem{chen2021deep}
Chun-Fu~Richard Chen, Rameswar Panda, Kandan Ramakrishnan, Rogerio Feris, John
  Cohn, Aude Oliva, and Quanfu Fan.
\newblock Deep analysis of cnn-based spatio-temporal representations for action
  recognition.
\newblock In {\em Proceedings of the IEEE/CVF Conference on Computer Vision and
  Pattern Recognition}, pages 6165--6175, 2021.

\bibitem{chen2018optimizing}
Kai Chen, Jiaqi Wang, Shuo Yang, Xingcheng Zhang, Yuanjun Xiong, Chen~Change
  Loy, and Dahua Lin.
\newblock Optimizing video object detection via a scale-time lattice.
\newblock In {\em Proceedings of the IEEE conference on computer vision and
  pattern recognition}, pages 7814--7823, 2018.

\bibitem{chen2020memory}
Yihong Chen, Yue Cao, Han Hu, and Liwei Wang.
\newblock Memory enhanced global-local aggregation for video object detection.
\newblock In {\em Proceedings of the IEEE/CVF Conference on Computer Vision and
  Pattern Recognition}, pages 10337--10346, 2020.

\bibitem{cheron2015p}
Guilhem Ch{\'e}ron, Ivan Laptev, and Cordelia Schmid.
\newblock P-cnn: Pose-based cnn features for action recognition.
\newblock In {\em ICCV-IEEE International Conference on Computer Vision}, pages
  3218--3226. IEEE, 2015.

\bibitem{crasto2019mars}
Nieves Crasto, Philippe Weinzaepfel, Karteek Alahari, and Cordelia Schmid.
\newblock Mars: Motion-augmented rgb stream for action recognition.
\newblock In {\em Proceedings of the IEEE/CVF Conference on Computer Vision and
  Pattern Recognition}, pages 7882--7891, 2019.

\bibitem{cui2021tf}
Yiming Cui, Liqi Yan, Zhiwen Cao, and Dongfang Liu.
\newblock Tf-blender: Temporal feature blender for video object detection.
\newblock In {\em Proceedings of the IEEE/CVF International Conference on
  Computer Vision}, pages 8138--8147, 2021.

\bibitem{deng2019object}
Hanming Deng, Yang Hua, Tao Song, Zongpu Zhang, Zhengui Xue, Ruhui Ma, Neil
  Robertson, and Haibing Guan.
\newblock Object guided external memory network for video object detection.
\newblock In {\em Proceedings of the IEEE/CVF International Conference on
  Computer Vision}, pages 6678--6687, 2019.

\bibitem{deng2019relation}
Jiajun Deng, Yingwei Pan, Ting Yao, Wengang Zhou, Houqiang Li, and Tao Mei.
\newblock Relation distillation networks for video object detection.
\newblock In {\em Proceedings of the IEEE/CVF International Conference on
  Computer Vision}, pages 7023--7032, 2019.

\bibitem{diba2018spatio}
Ali Diba, Mohsen Fayyaz, Vivek Sharma, M~Mahdi Arzani, Rahman Yousefzadeh,
  Juergen Gall, and Luc Van~Gool.
\newblock Spatio-temporal channel correlation networks for action
  classification.
\newblock In {\em Proceedings of the European Conference on Computer Vision
  (ECCV)}, pages 284--299, 2018.

\bibitem{everingham2010pascal}
Mark Everingham, Luc Van~Gool, Christopher~KI Williams, John Winn, and Andrew
  Zisserman.
\newblock The pascal visual object classes (voc) challenge.
\newblock {\em International journal of computer vision}, 88(2):303--338, 2010.

\bibitem{fan2019more}
Quanfu Fan, Chun-Fu Chen, Hilde Kuehne, Marco Pistoia, and David Cox.
\newblock More is less: Learning efficient video representations by big-little
  network and depthwise temporal aggregation.
\newblock {\em arXiv preprint arXiv:1912.00869}, 2019.

\bibitem{feichtenhofer2017detect}
Christoph Feichtenhofer, Axel Pinz, and Andrew Zisserman.
\newblock Detect to track and track to detect.
\newblock In {\em Proceedings of the IEEE International Conference on Computer
  Vision}, pages 3038--3046, 2017.

\bibitem{geng2020object}
Qichuan Geng, Hong Zhang, Na Jiang, Xiaojuan Qi, Liangjun Zhang, and Zhong
  Zhou.
\newblock Object-aware feature aggregation for video object detection.
\newblock {\em arXiv preprint arXiv:2010.12573}, 2020.

\bibitem{girshick2015fast}
Ross Girshick.
\newblock Fast r-cnn.
\newblock In {\em Proceedings of the IEEE international conference on computer
  vision}, pages 1440--1448, 2015.

\bibitem{girshick2015region}
Ross Girshick, Jeff Donahue, Trevor Darrell, and Jitendra Malik.
\newblock Region-based convolutional networks for accurate object detection and
  segmentation.
\newblock {\em IEEE transactions on pattern analysis and machine intelligence},
  38(1):142--158, 2015.

\bibitem{goyal2017something}
Raghav Goyal, Samira Ebrahimi~Kahou, Vincent Michalski, Joanna Materzynska,
  Susanne Westphal, Heuna Kim, Valentin Haenel, Ingo Fruend, Peter Yianilos,
  Moritz Mueller-Freitag, et~al.
\newblock The" something something" video database for learning and evaluating
  visual common sense.
\newblock In {\em Proceedings of the IEEE international conference on computer
  vision}, pages 5842--5850, 2017.

\bibitem{gu2018ava}
Chunhui Gu, Chen Sun, David~A Ross, Carl Vondrick, Caroline Pantofaru, Yeqing
  Li, Sudheendra Vijayanarasimhan, George Toderici, Susanna Ricco, Rahul
  Sukthankar, et~al.
\newblock Ava: A video dataset of spatio-temporally localized atomic visual
  actions.
\newblock In {\em Proceedings of the IEEE Conference on Computer Vision and
  Pattern Recognition}, pages 6047--6056, 2018.

\bibitem{guo2019progressive}
Chaoxu Guo, Bin Fan, Jie Gu, Qian Zhang, Shiming Xiang, Veronique Prinet, and
  Chunhong Pan.
\newblock Progressive sparse local attention for video object detection.
\newblock In {\em Proceedings of the IEEE/CVF International Conference on
  Computer Vision}, pages 3909--3918, 2019.

\bibitem{hagen2013video}
Nathan Hagen, Robert~T Kester, Christopher~G Morlier, Jeffrey~A Panek, Paul
  Drayton, Dave Fashimpaur, Paul Stone, and Elizabeth Adams.
\newblock Video-rate spectral imaging of gas leaks in the longwave infrared.
\newblock In {\em Chemical, Biological, Radiological, Nuclear, and Explosives
  (CBRNE) Sensing XIV}, volume 8710, page 871005. International Society for
  Optics and Photonics, 2013.

\bibitem{han2020mining}
Mingfei Han, Yali Wang, Xiaojun Chang, and Yu Qiao.
\newblock Mining inter-video proposal relations for video object detection.
\newblock In {\em European Conference on Computer Vision}, pages 431--446.
  Springer, 2020.

\bibitem{han2016seq}
Wei Han, Pooya Khorrami, Tom~Le Paine, Prajit Ramachandran, Mohammad
  Babaeizadeh, Honghui Shi, Jianan Li, Shuicheng Yan, and Thomas~S Huang.
\newblock Seq-nms for video object detection.
\newblock {\em arXiv preprint arXiv:1602.08465}, 2016.

\bibitem{hara2018can}
Kensho Hara, Hirokatsu Kataoka, and Yutaka Satoh.
\newblock Can spatiotemporal 3d cnns retrace the history of 2d cnns and
  imagenet?
\newblock In {\em Proceedings of the IEEE conference on Computer Vision and
  Pattern Recognition}, pages 6546--6555, 2018.

\bibitem{he2016deep}
Kaiming He, Xiangyu Zhang, Shaoqing Ren, and Jian Sun.
\newblock Deep residual learning for image recognition.
\newblock In {\em Proceedings of the IEEE conference on computer vision and
  pattern recognition}, pages 770--778, 2016.

\bibitem{he2021end}
Lu He, Qianyu Zhou, Xiangtai Li, Li Niu, Guangliang Cheng, Xiao Li, Wenxuan
  Liu, Yunhai Tong, Lizhuang Ma, and Liqing Zhang.
\newblock End-to-end video object detection with spatial-temporal transformers.
\newblock {\em arXiv preprint arXiv:2105.10920}, 2021.

\bibitem{hou2017tube}
Rui Hou, Chen Chen, and Mubarak Shah.
\newblock Tube convolutional neural network (t-cnn) for action detection in
  videos.
\newblock In {\em Proceedings of the IEEE international conference on computer
  vision}, pages 5822--5831, 2017.

\bibitem{hou2007saliency}
Xiaodi Hou and Liqing Zhang.
\newblock Saliency detection: A spectral residual approach.
\newblock In {\em 2007 IEEE Conference on computer vision and pattern
  recognition}, pages 1--8. Ieee, 2007.

\bibitem{hwang2015multispectral}
Soonmin Hwang, Jaesik Park, Namil Kim, Yukyung Choi, and In So~Kweon.
\newblock Multispectral pedestrian detection: Benchmark dataset and baseline.
\newblock In {\em Proceedings of the IEEE conference on computer vision and
  pattern recognition}, pages 1037--1045, 2015.

\bibitem{jhuang2013towards}
Hueihan Jhuang, Juergen Gall, Silvia Zuffi, Cordelia Schmid, and Michael~J
  Black.
\newblock Towards understanding action recognition.
\newblock In {\em Proceedings of the IEEE international conference on computer
  vision}, pages 3192--3199, 2013.

\bibitem{jiang2019stm}
Boyuan Jiang, MengMeng Wang, Weihao Gan, Wei Wu, and Junjie Yan.
\newblock Stm: Spatiotemporal and motion encoding for action recognition.
\newblock In {\em Proceedings of the IEEE/CVF International Conference on
  Computer Vision}, pages 2000--2009, 2019.

\bibitem{jiang2020learning}
Zhengkai Jiang, Yu Liu, Ceyuan Yang, Jihao Liu, Peng Gao, Qian Zhang, Shiming
  Xiang, and Chunhong Pan.
\newblock Learning where to focus for efficient video object detection.
\newblock In {\em European Conference on Computer Vision}, pages 18--34.
  Springer, 2020.

\bibitem{jiao2021new}
Licheng Jiao, Ruohan Zhang, Fang Liu, Shuyuan Yang, Biao Hou, Lingling Li, and
  Xu Tang.
\newblock New generation deep learning for video object detection: A survey.
\newblock {\em IEEE Transactions on Neural Networks and Learning Systems},
  2021.

\bibitem{jin2022feature}
Ruibing Jin, Guosheng Lin, Changyun Wen, Jianliang Wang, and Fayao Liu.
\newblock Feature flow: In-network feature flow estimation for video object
  detection.
\newblock {\em Pattern Recognition}, 122:108323, 2022.

\bibitem{kalogeiton2017action}
Vicky Kalogeiton, Philippe Weinzaepfel, Vittorio Ferrari, and Cordelia Schmid.
\newblock Action tubelet detector for spatio-temporal action localization.
\newblock In {\em Proceedings of the IEEE International Conference on Computer
  Vision}, pages 4405--4413, 2017.

\bibitem{kang2017t}
Kai Kang, Hongsheng Li, Junjie Yan, Xingyu Zeng, Bin Yang, Tong Xiao, Cong
  Zhang, Zhe Wang, Ruohui Wang, Xiaogang Wang, et~al.
\newblock T-cnn: Tubelets with convolutional neural networks for object
  detection from videos.
\newblock {\em IEEE Transactions on Circuits and Systems for Video Technology},
  28(10):2896--2907, 2017.

\bibitem{koh2021joint}
Junho Koh, Jaekyum Kim, Younji Shin, Byeongwon Lee, Seungji Yang, and Jun~Won
  Choi.
\newblock Joint representation of temporal image sequences and object motion
  for video object detection.
\newblock In {\em 2021 IEEE International Conference on Robotics and Automation
  (ICRA)}, pages 13370--13376. IEEE, 2021.

\bibitem{kwon2020motionsqueeze}
Heeseung Kwon, Manjin Kim, Suha Kwak, and Minsu Cho.
\newblock Motionsqueeze: Neural motion feature learning for video
  understanding.
\newblock In {\em European Conference on Computer Vision}, pages 345--362.
  Springer, 2020.

\bibitem{li2021time}
Changhai Li, Huawei Chen, Jingqing Lu, Yang Huang, and Yingying Liu.
\newblock Time and frequency network for human action detection in videos.
\newblock {\em arXiv preprint arXiv:2103.04680}, 2021.

\bibitem{li2018recurrent}
Dong Li, Zhaofan Qiu, Qi Dai, Ting Yao, and Tao Mei.
\newblock Recurrent tubelet proposal and recognition networks for action
  detection.
\newblock In {\em Proceedings of the European conference on computer vision
  (ECCV)}, pages 303--318, 2018.

\bibitem{li2020tea}
Yan Li, Bin Ji, Xintian Shi, Jianguo Zhang, Bin Kang, and Limin Wang.
\newblock Tea: Temporal excitation and aggregation for action recognition.
\newblock In {\em Proceedings of the IEEE/CVF Conference on Computer Vision and
  Pattern Recognition}, pages 909--918, 2020.

\bibitem{li2020actions}
Yixuan Li, Zixu Wang, Limin Wang, and Gangshan Wu.
\newblock Actions as moving points.
\newblock In {\em European Conference on Computer Vision}, pages 68--84.
  Springer, 2020.

\bibitem{lin2019tsm}
Ji Lin, Chuang Gan, and Song Han.
\newblock Tsm: Temporal shift module for efficient video understanding.
\newblock In {\em Proceedings of the IEEE/CVF International Conference on
  Computer Vision}, pages 7083--7093, 2019.

\bibitem{lin2014microsoft}
Tsung-Yi Lin, Michael Maire, Serge Belongie, James Hays, Pietro Perona, Deva
  Ramanan, Piotr Doll{\'a}r, and C~Lawrence Zitnick.
\newblock Microsoft coco: Common objects in context.
\newblock In {\em European conference on computer vision}, pages 740--755.
  Springer, 2014.

\bibitem{liu2016ssd}
Wei Liu, Dragomir Anguelov, Dumitru Erhan, Christian Szegedy, Scott Reed,
  Cheng-Yang Fu, and Alexander~C Berg.
\newblock Ssd: Single shot multibox detector.
\newblock In {\em European conference on computer vision}, pages 21--37.
  Springer, 2016.

\bibitem{liu2021acdnet}
Yu Liu, Fan Yang, and Dominique Ginhac.
\newblock Acdnet: An action detection network for real-time edge computing
  based on flow-guided feature approximation and memory aggregation.
\newblock {\em Pattern Recognition Letters}, 145:118--126, 2021.

\bibitem{liu2020teinet}
Zhaoyang Liu, Donghao Luo, Yabiao Wang, Limin Wang, Ying Tai, Chengjie Wang,
  Jilin Li, Feiyue Huang, and Tong Lu.
\newblock Teinet: Towards an efficient architecture for video recognition.
\newblock In {\em Proceedings of the AAAI Conference on Artificial
  Intelligence}, volume~34, pages 11669--11676, 2020.

\bibitem{luo2019object}
Hao Luo, Lichao Huang, Han Shen, Yuan Li, Chang Huang, and Xinggang Wang.
\newblock Object detection in video with spatial-temporal context aggregation.
\newblock {\em arXiv preprint arXiv:1907.04988}, 2019.

\bibitem{mei2021sanet}
Jie Mei, Ming-Ming Cheng, Gang Xu, Lan-Ruo Wan, and Huan Zhang.
\newblock Sanet: A slice-aware network for pulmonary nodule detection.
\newblock {\em IEEE transactions on pattern analysis and machine intelligence},
  2021.

\bibitem{pan2021actor}
Junting Pan, Siyu Chen, Mike~Zheng Shou, Yu Liu, Jing Shao, and Hongsheng Li.
\newblock Actor-context-actor relation network for spatio-temporal action
  localization.
\newblock In {\em Proceedings of the IEEE/CVF Conference on Computer Vision and
  Pattern Recognition}, pages 464--474, 2021.

\bibitem{peng2016multi}
Xiaojiang Peng and Cordelia Schmid.
\newblock Multi-region two-stream r-cnn for action detection.
\newblock In {\em European conference on computer vision}, pages 744--759.
  Springer, 2016.

\bibitem{qiu2017learning}
Zhaofan Qiu, Ting Yao, and Tao Mei.
\newblock Learning spatio-temporal representation with pseudo-3d residual
  networks.
\newblock In {\em proceedings of the IEEE International Conference on Computer
  Vision}, pages 5533--5541, 2017.

\bibitem{ren2015faster}
Shaoqing Ren, Kaiming He, Ross Girshick, and Jian Sun.
\newblock Faster r-cnn: Towards real-time object detection with region proposal
  networks.
\newblock In {\em Advances in neural information processing systems}, pages
  91--99, 2015.

\bibitem{russakovsky2015imagenet}
Olga Russakovsky, Jia Deng, Hao Su, Jonathan Krause, Sanjeev Satheesh, Sean Ma,
  Zhiheng Huang, Andrej Karpathy, Aditya Khosla, Michael Bernstein, et~al.
\newblock Imagenet large scale visual recognition challenge.
\newblock {\em International journal of computer vision}, 115(3):211--252,
  2015.

\bibitem{sarmiento2021spatio}
Manuel Sarmiento~Calder{\'o}, David Varas, and Elisenda Bou-Balust.
\newblock Spatio-temporal context for action detection.
\newblock {\em arXiv e-prints}, pages arXiv--2106, 2021.

\bibitem{shao2020temporal}
Hao Shao, Shengju Qian, and Yu Liu.
\newblock Temporal interlacing network.
\newblock In {\em Proceedings of the AAAI Conference on Artificial
  Intelligence}, volume~34, pages 11966--11973, 2020.

\bibitem{shvets2019leveraging}
Mykhailo Shvets, Wei Liu, and Alexander~C Berg.
\newblock Leveraging long-range temporal relationships between proposals for
  video object detection.
\newblock In {\em Proceedings of the IEEE/CVF International Conference on
  Computer Vision}, pages 9756--9764, 2019.

\bibitem{soomro2012ucf101}
Khurram Soomro, Amir~Roshan Zamir, and Mubarak Shah.
\newblock Ucf101: A dataset of 101 human actions classes from videos in the
  wild.
\newblock {\em arXiv preprint arXiv:1212.0402}, 2012.

\bibitem{stauffer1999adaptive}
Chris Stauffer and W~Eric~L Grimson.
\newblock Adaptive background mixture models for real-time tracking.
\newblock In {\em Proceedings. 1999 IEEE computer society conference on
  computer vision and pattern recognition (Cat. No PR00149)}, volume~2, pages
  246--252. IEEE, 1999.

\bibitem{stroud2020d3d}
Jonathan Stroud, David Ross, Chen Sun, Jia Deng, and Rahul Sukthankar.
\newblock D3d: Distilled 3d networks for video action recognition.
\newblock In {\em Proceedings of the IEEE/CVF Winter Conference on Applications
  of Computer Vision}, pages 625--634, 2020.

\bibitem{sun2018actor}
Chen Sun, Abhinav Shrivastava, Carl Vondrick, Kevin Murphy, Rahul Sukthankar,
  and Cordelia Schmid.
\newblock Actor-centric relation network.
\newblock In {\em Proceedings of the European Conference on Computer Vision
  (ECCV)}, pages 318--334, 2018.

\bibitem{tran2015learning}
Du Tran, Lubomir Bourdev, Rob Fergus, Lorenzo Torresani, and Manohar Paluri.
\newblock Learning spatiotemporal features with 3d convolutional networks.
\newblock In {\em Proceedings of the IEEE international conference on computer
  vision}, pages 4489--4497, 2015.

\bibitem{tran2018closer}
Du Tran, Heng Wang, Lorenzo Torresani, Jamie Ray, Yann LeCun, and Manohar
  Paluri.
\newblock A closer look at spatiotemporal convolutions for action recognition.
\newblock In {\em Proceedings of the IEEE conference on Computer Vision and
  Pattern Recognition}, pages 6450--6459, 2018.

\bibitem{wang2016actionness}
Limin Wang, Yu Qiao, Xiaoou Tang, and Luc Van~Gool.
\newblock Actionness estimation using hybrid fully convolutional networks.
\newblock In {\em Proceedings of the IEEE Conference on Computer Vision and
  Pattern Recognition}, pages 2708--2717, 2016.

\bibitem{wang2021tdn}
Limin Wang, Zhan Tong, Bin Ji, and Gangshan Wu.
\newblock Tdn: Temporal difference networks for efficient action recognition.
\newblock In {\em Proceedings of the IEEE/CVF Conference on Computer Vision and
  Pattern Recognition}, pages 1895--1904, 2021.

\bibitem{wang2016temporal}
Limin Wang, Yuanjun Xiong, Zhe Wang, Yu Qiao, Dahua Lin, Xiaoou Tang, and Luc
  Van~Gool.
\newblock Temporal segment networks: Towards good practices for deep action
  recognition.
\newblock In {\em European conference on computer vision}, pages 20--36.
  Springer, 2016.

\bibitem{wang2018fully}
Shiyao Wang, Yucong Zhou, Junjie Yan, and Zhidong Deng.
\newblock Fully motion-aware network for video object detection.
\newblock In {\em Proceedings of the European conference on computer vision
  (ECCV)}, pages 542--557, 2018.

\bibitem{wang2021real}
Xinggang Wang, Zhaojin Huang, Bencheng Liao, Lichao Huang, Yongchao Gong, and
  Chang Huang.
\newblock Real-time and accurate object detection in compressed video by long
  short-term feature aggregation.
\newblock {\em Computer Vision and Image Understanding}, 206:103188, 2021.

\bibitem{wu2019long}
Chao-Yuan Wu, Christoph Feichtenhofer, Haoqi Fan, Kaiming He, Philipp
  Krahenbuhl, and Ross Girshick.
\newblock Long-term feature banks for detailed video understanding.
\newblock In {\em Proceedings of the IEEE/CVF Conference on Computer Vision and
  Pattern Recognition}, pages 284--293, 2019.

\bibitem{wu2019sequence}
Haiping Wu, Yuntao Chen, Naiyan Wang, and Zhaoxiang Zhang.
\newblock Sequence level semantics aggregation for video object detection.
\newblock In {\em Proceedings of the IEEE/CVF International Conference on
  Computer Vision}, pages 9217--9225, 2019.

\bibitem{wu2020context}
Jianchao Wu, Zhanghui Kuang, Limin Wang, Wayne Zhang, and Gangshan Wu.
\newblock Context-aware rcnn: A baseline for action detection in videos.
\newblock In {\em European Conference on Computer Vision}, pages 440--456.
  Springer, 2020.

\bibitem{xiao2018video}
Fanyi Xiao and Yong~Jae Lee.
\newblock Video object detection with an aligned spatial-temporal memory.
\newblock In {\em Proceedings of the European Conference on Computer Vision
  (ECCV)}, pages 485--501, 2018.

\bibitem{xie2018rethinking}
Saining Xie, Chen Sun, Jonathan Huang, Zhuowen Tu, and Kevin Murphy.
\newblock Rethinking spatiotemporal feature learning: Speed-accuracy trade-offs
  in video classification.
\newblock In {\em Proceedings of the European conference on computer vision
  (ECCV)}, pages 305--321, 2018.

\bibitem{yang2019step}
Xitong Yang, Xiaodong Yang, Ming-Yu Liu, Fanyi Xiao, Larry~S Davis, and Jan
  Kautz.
\newblock Step: Spatio-temporal progressive learning for video action
  detection.
\newblock In {\em Proceedings of the IEEE/CVF Conference on Computer Vision and
  Pattern Recognition}, pages 264--272, 2019.

\bibitem{yao2020video}
Chun-Han Yao, Chen Fang, Xiaohui Shen, Yangyue Wan, and Ming-Hsuan Yang.
\newblock Video object detection via object-level temporal aggregation.
\newblock In {\em European conference on computer vision}, pages 160--177.
  Springer, 2020.

\bibitem{yu2018deep}
Fisher Yu, Dequan Wang, Evan Shelhamer, and Trevor Darrell.
\newblock Deep layer aggregation.
\newblock In {\em Proceedings of the IEEE conference on computer vision and
  pattern recognition}, pages 2403--2412, 2018.

\bibitem{zhang2020pan}
Can Zhang, Yuexian Zou, Guang Chen, and Lei Gan.
\newblock Pan: Towards fast action recognition via learning persistence of
  appearance.
\newblock {\em arXiv preprint arXiv:2008.03462}, 2020.

\bibitem{zhang2020hyperli}
Tianwen Zhang, Xiaoling Zhang, Jun Shi, and Shunjun Wei.
\newblock Hyperli-net: A hyper-light deep learning network for high-accurate
  and high-speed ship detection from synthetic aperture radar imagery.
\newblock {\em ISPRS Journal of Photogrammetry and Remote Sensing},
  167:123--153, 2020.

\bibitem{zhao2018recognize}
Yue Zhao, Yuanjun Xiong, and Dahua Lin.
\newblock Recognize actions by disentangling components of dynamics.
\newblock In {\em Proceedings of the IEEE Conference on Computer Vision and
  Pattern Recognition}, pages 6566--6575, 2018.

\bibitem{zheng2020distance}
Zhaohui Zheng, Ping Wang, Wei Liu, Jinze Li, Rongguang Ye, and Dongwei Ren.
\newblock Distance-iou loss: Faster and better learning for bounding box
  regression.
\newblock In {\em Proceedings of the AAAI Conference on Artificial
  Intelligence}, volume~34, pages 12993--13000, 2020.

\bibitem{zheng2021localization}
Zhaohui Zheng, Rongguang Ye, Ping Wang, Jun Wang, Dongwei Ren, and Wangmeng
  Zuo.
\newblock Localization distillation for object detection.
\newblock {\em arXiv preprint arXiv:2102.12252}, 2021.

\bibitem{zhou2019objects}
Xingyi Zhou, Dequan Wang, and Philipp Kr{\"a}henb{\"u}hl.
\newblock Objects as points.
\newblock {\em arXiv preprint arXiv:1904.07850}, 2019.

\bibitem{zhu2018towards}
Xizhou Zhu, Jifeng Dai, Lu Yuan, and Yichen Wei.
\newblock Towards high performance video object detection.
\newblock In {\em Proceedings of the IEEE Conference on Computer Vision and
  Pattern Recognition}, pages 7210--7218, 2018.

\bibitem{zhu2017flow}
Xizhou Zhu, Yujie Wang, Jifeng Dai, Lu Yuan, and Yichen Wei.
\newblock Flow-guided feature aggregation for video object detection.
\newblock In {\em Proceedings of the IEEE International Conference on Computer
  Vision}, pages 408--417, 2017.

\bibitem{zhu2017deep}
Xizhou Zhu, Yuwen Xiong, Jifeng Dai, Lu Yuan, and Yichen Wei.
\newblock Deep feature flow for video recognition.
\newblock In {\em Proceedings of the IEEE conference on computer vision and
  pattern recognition}, pages 2349--2358, 2017.

\bibitem{zhu2020comprehensive}
Yi Zhu, Xinyu Li, Chunhui Liu, Mohammadreza Zolfaghari, Yuanjun Xiong, Chongruo
  Wu, Zhi Zhang, Joseph Tighe, R Manmatha, and Mu Li.
\newblock A comprehensive study of deep video action recognition.
\newblock {\em arXiv preprint arXiv:2012.06567}, 2020.

\end{thebibliography}
}



\begin{figure*}[!t]
	\begin{center}
		\includegraphics[width=1.0\linewidth]{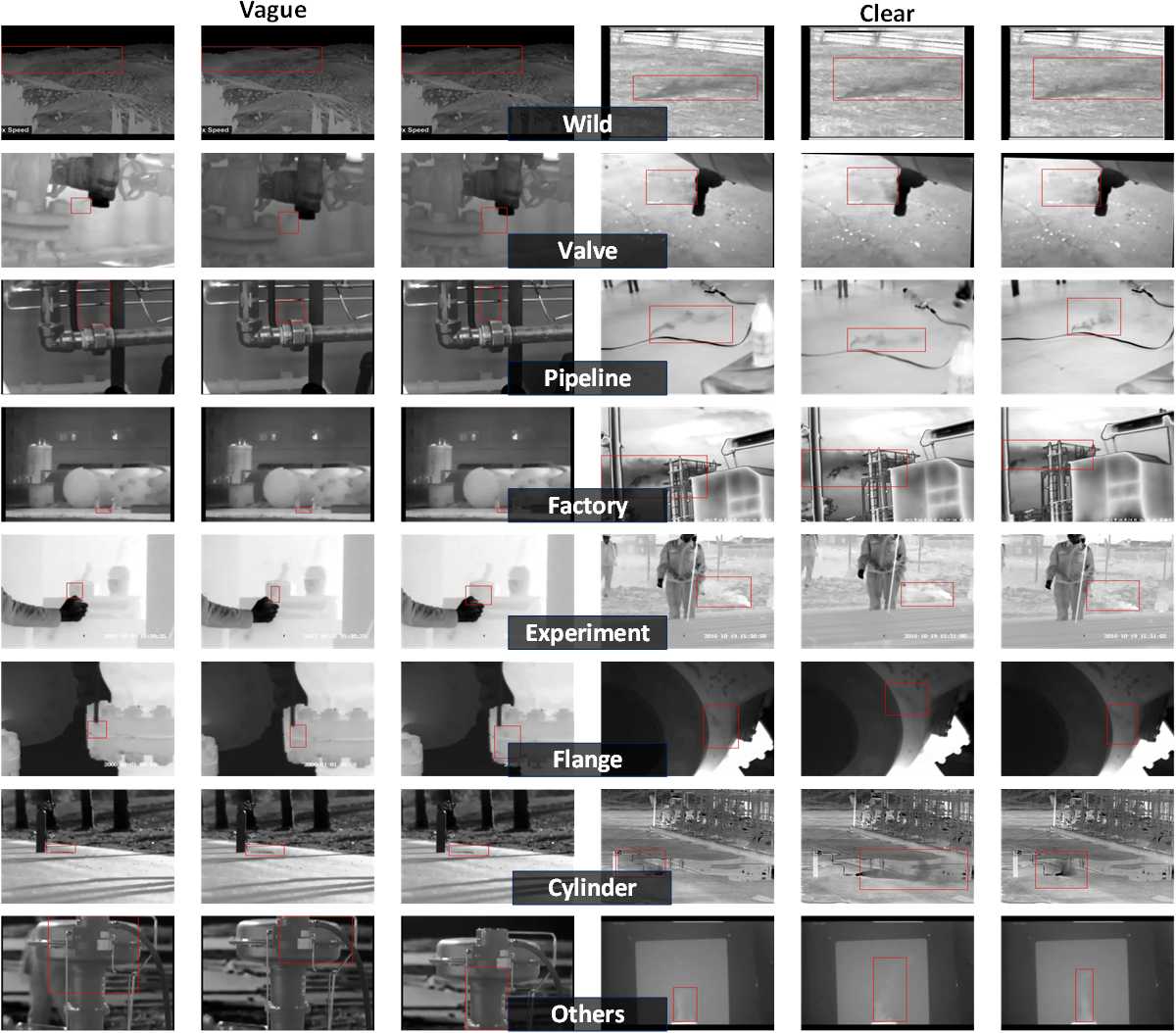}
	\end{center}
	\vspace{-1.5em}
	\caption{The overview of the IOD-Video Dataset, including the following scenes: wild, valve, pipeline, factory, experiment, flange, cylinder and others.}
	\label{fig:overview}
	\vspace{-1.0em}
\end{figure*}

\newpage
{
~~~~~~~~~~~~~\Large \textbf{Supplementary Material}
\\ 

}
\large\noindent\textbf{1. IOD-Video Dataset Overview}

\normalsize Fig.~\ref{fig:overview} shows the overview of the IOD-Video dataset. Some of the scenes are active gas leak of experimenters such as samples in the ``experiment" and ``cylinder" set. Some of the scenes are captured in the industrial environment and they are classified into ``pipeline", ``flange", ``valve" and ``factory" sets according to the background. In additional, ``wild" set is the natural scene and the rest are classified as ``others". The left side is the vague samples which are difficult for human eyes to judge the boundary of the object, while the clear samples in the right side have relatively clear boundaries. Above all, whether it is vague or clear sample, spatio-temporal feature aggregation is very important.

\begin{figure*}[t]
	\begin{center} 
		\includegraphics[width=1.0\linewidth]{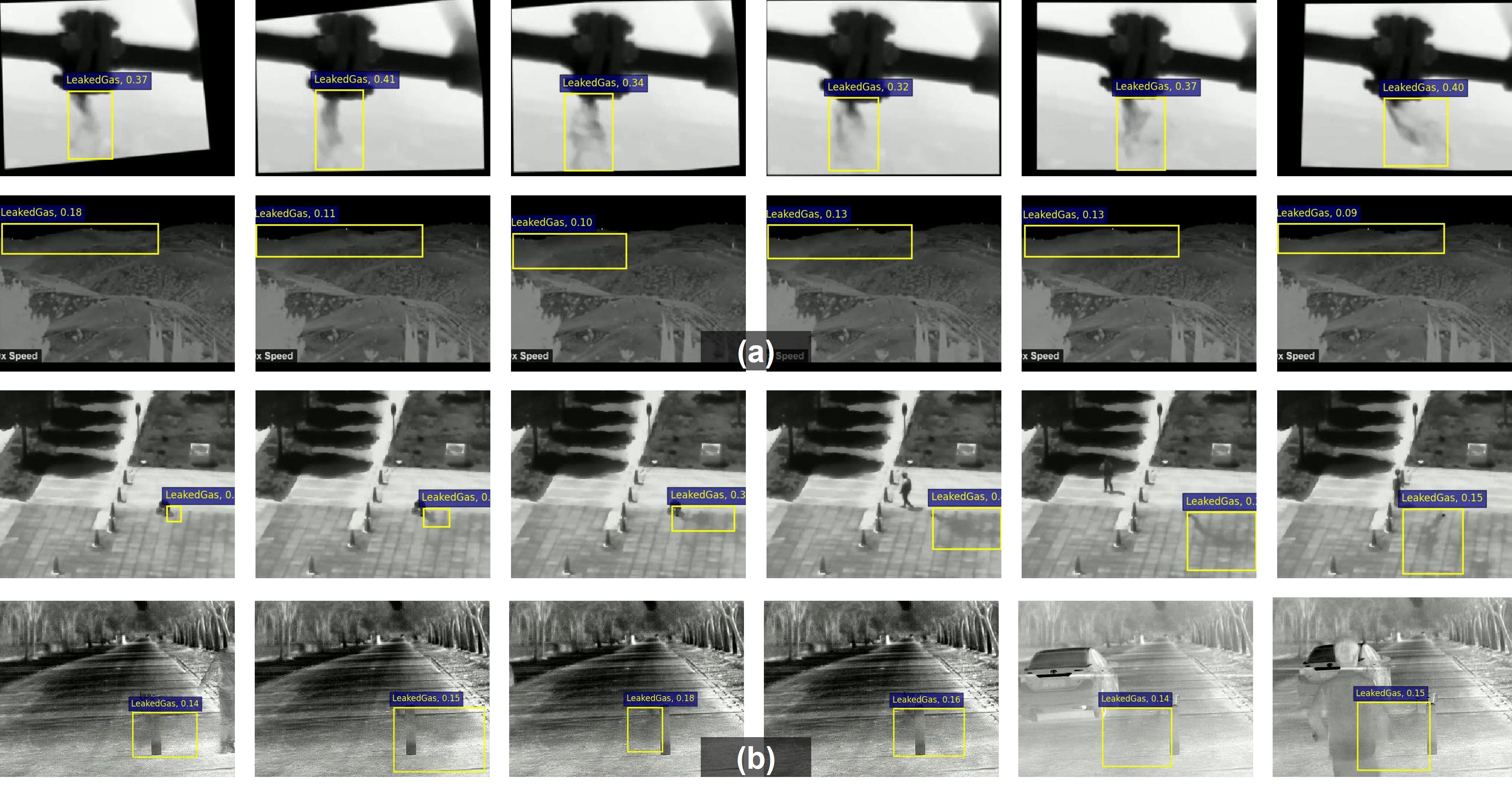}
	\end{center}
	\vspace{-1.5em}
	\caption{More visualization results of our baseline method (TEA+STAloss) on the test set of IOD-Video dataset (a) and open world (b) scenes.   }
	\label{fig:Visualization2}
	\vspace{-1.0em}
\end{figure*}

\large\noindent\textbf{2. More Visualization Results}

\normalsize Here shows more visualization results of our baseline method (TEA+STAloss). Fig.~\ref{fig:Visualization2} (a) is the representative samples of split-3 test set on the IOD-Video dataset. The successive images of the first row have dramatic change of morphologic, but our proposed method can track the object continuously and tighten the boundary well.  Although the sample in the second row varies little in the position and shape, we can hardly identity where the insubstantial object is. Fig.~\ref{fig:Visualization2} (b) shows two video samples in the open world which utilize the cylinder to simulate gas leaks. They both presents bad visual saliency and dramatic shape variation. Nevertheless, with spatio-temporal aggregation design in the backbone and loss level, our baseline method can alleviate the problem of information deficiencies in a single frame. It is robust to the human body occlusion and surpasses the capability of human eyes in some certain scenes.

\large\noindent\textbf{3. Limitation of STAloss}

\begin{figure}[t]
	\centering
	\includegraphics[width=1.0\linewidth]{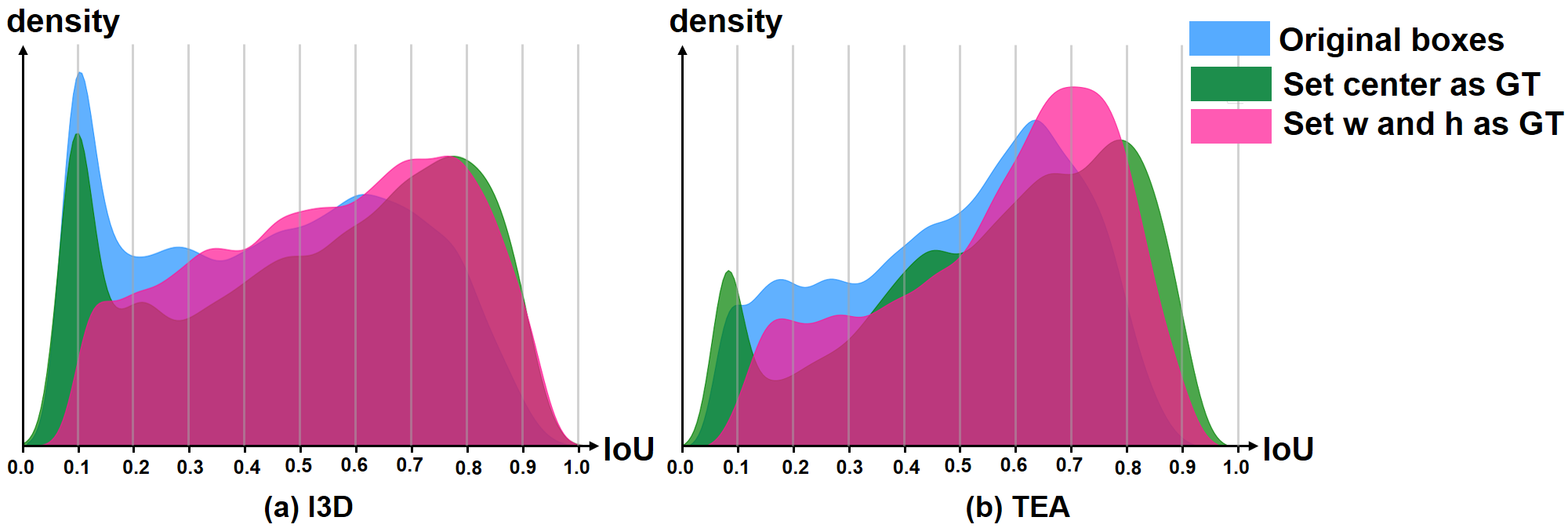}
	\vspace{-1.8em}
	\caption{The statistic of IoU density distribution on the test set of split-1. The blue area is the IoU between the original predicted boxes and GT boxes. The green area sets the center of predicted boxes as GT while remains the width (w) and height (h) as original prediction. The pink area sets the width and height as GT while remains the center as original prediction. }
	\label{density}
	\vspace{-0.5em}
\end{figure}

\normalsize 
 Although STAloss has achieved good results on the TEA backbone, Tab.~\ref{STAlosseffect} shows it fails on the I3D backbone. To explore the degradation reason, we make statistics of the IoU distribution between the predicted boxes and GT boxes on the split-1 test set. The blue color area shows the IoU distribution of original predicted boxes with GT boxes. On the one hand, compared with TEA, I3D is more distributed in the region with low IoU, and the accuracy of the predicted boxes obtained by I3D is relatively low, it is difficult for STAloss to optimize on the I3D backbone. On the other hand, we set the center of the prediction box as GT, and keep the width and height as the original prediction result, which is the red area in Fig.~\ref{density}. Meanwhile, we set the width and height of the prediction box as GT, and keep the center as the original prediction result, which is the pink area in Fig.~\ref{density}. It can be seen that the pink area covers higher IOU regions, which indicates that our baseline model suffers more from the inaccurate prediction of the width and height, while STAloss only makes adjustment of the center which limits the performance gain.
 
\renewcommand{\arraystretch}{1.2}
\begin{table}[]
	\begin{center}
		\setlength{\tabcolsep}{1mm}{
			\begin{tabular}{c|ccccc}
				\hline
				\multirow{2}{*}{\begin{tabular}[c]{@{}c@{}} \\ \end{tabular}} & \multicolumn{5}{c}{Frame AP (\%)}        \\ \cline{2-6} 
				& @0.5 & @0.75 & @clear & @vague & 0.5:0.95 \\ \hline
				I3D                                                                      & 36.83 & 7.39   & 18.78  & 9.29 & 13.43   \\
				I3D+STAloss                                                                        & 36.17 &  6.80  &18.05  & 9.19   & 12.98     \\
				TEA                                                                        & 42.19 & 8.66 & 22.97  &9.95 & 15.69    \\
				TEA+STAloss                                                                         & 45.08 & 9.50 & 24.43  & 10.91   & 16.99     \\ \hline
		\end{tabular}}
		\end {center}
		\vspace{-0.6cm} 
		\caption{Evaluation results of the STAloss effect on I3D and TEA.}
		\label{STAlosseffect}
		\vspace{-0.2cm} 
	\end{table}

\begin{figure*}[t]
	\centering
	\includegraphics[width=1.0\linewidth]{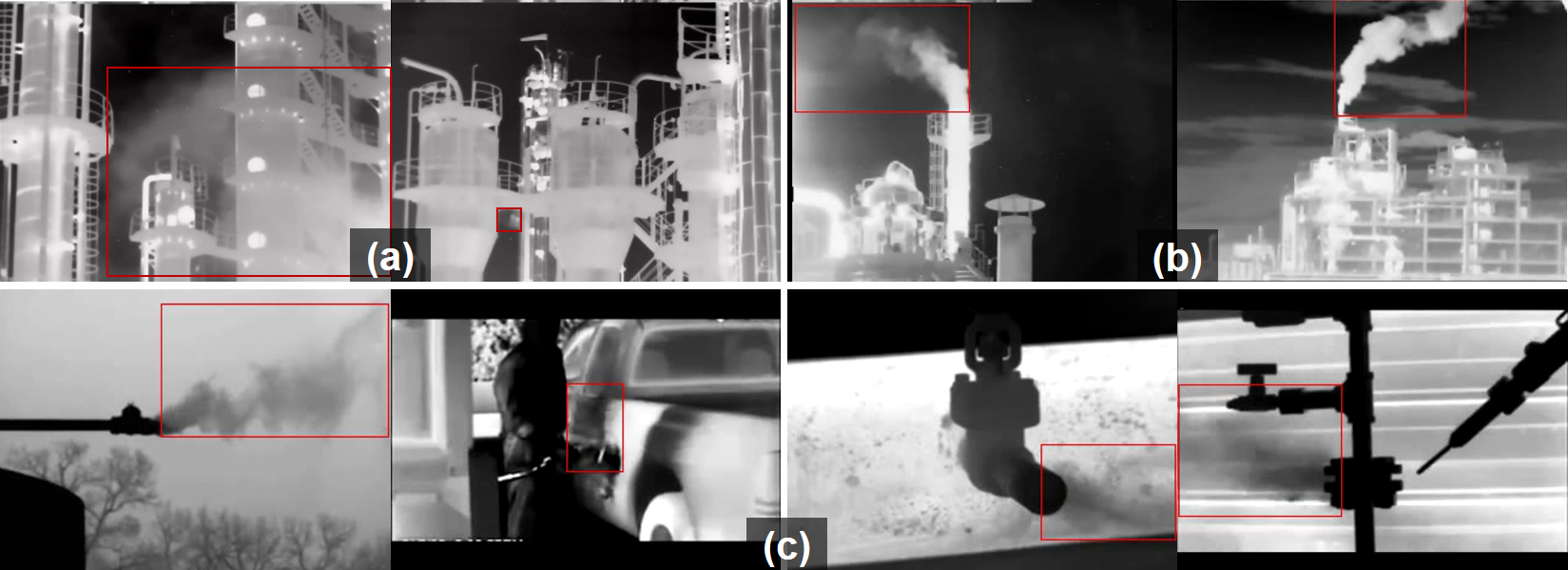}
	\vspace{-1.8em}
	\caption{ Some industrial applications of IOD task: (a) steam detection (b) industrial smoke monitoring (c) VOC gas leak detection. }
	\label{application2}
	\vspace{-0.5em}
\end{figure*}

\large\noindent\textbf{4. Industrial Applications}

\normalsize\noindent\textbf{Steam Detection.} Boiler tube and pipeline leakage is occurring frequently while traditional contact type sensors cannot operate in high temperature area, the camera-based solutions provide the advantages of wide area surveillance and long distance monitoring. The algorithm of IOD task can be applied to steam leakage in energy industry as shown in Fig.~\ref{application2} (a).

\noindent\textbf{Industrial Smoke Monitoring.} In the context of global carbon neutrality, environmental protection has become a problem that needs to be solved for the development of various industries. Industrial smoke emissions pose a serious threat to natural ecosystems which need to be monitored and regulated. The combination of computer vision technology and computational photography can provide a feasible solution since industrial smoke is clearer under mid-wave and long-wave infrared spectrum as shown in Fig.~\ref{application2} (b), which makes it possible for enterprises to improve their energy programs to meet emission requirements by enabling targeted maintenance and reduction.

\noindent\textbf{Volatile Organic Compound (VOC) Gas Leak Detection.} In the petroleum industry VOC gas leaks pose a significant risk at process facilities including fires, explosions and acute exposure to toxic gases. The accidental VOC gas leaks will lead to multiple fatalities or injuries, as well as loss of infrastructure which is critical to the business and economy. Nevertheless, most VOCs are invisible to human eyes, e.g. olefins and alkanes. They have a unique spectral characteristic and can be captured under specific infrared spectrum by multispectral cameras as shown in Fig.~\ref{application2} (c). The exploration of IOD task can promote the real-time intelligent monitoring for pinpointing the gas leak source and providing fast and accurate response, so as to generate an alarm well before the concentration reaches its lower explosive level (LEL) value and achieve the purpose of preventing consequences of accidents.

\end{document}